\documentclass{article}

\usepackage{microtype}
\usepackage{graphicx}
\usepackage{subfigure}
\usepackage{booktabs} 

\usepackage{graphicx} 
\usepackage{amsthm} 
\usepackage{pifont}

\usepackage{array}
\usepackage{url}            
\usepackage{amsfonts}       
\usepackage{nicefrac}       
\usepackage{microtype}      

\usepackage{amsmath} 
\usepackage{bm}

\usepackage{ulem}
\useunder{\uline}{\ul}{}
\usepackage{CJKulem}
\usepackage{colortbl}
\usepackage[table]{xcolor}
\definecolor{green}{RGB}{199, 223, 181}
\definecolor{orange}{RGB}{243, 204, 174}
\definecolor{blue1}{RGB}{193, 214, 237}

\usepackage{multirow}
\usepackage{longtable}
\usepackage{bbding} %

\let\emph\relax
\DeclareTextFontCommand{\emph}{\rm}
\usepackage{hyperref}
\usepackage[accepted]{icml2025}

\usepackage{amsmath}
\usepackage{amssymb}
\usepackage{mathtools}
\usepackage{amsthm}
\usepackage{tcolorbox}
\tcbset{
  colframe=teal,        
  colback=green!50,      
  boxrule=0.3mm,        
  arc=3mm,              
  left=3mm,             
  right=3mm,            
  top=2mm,              
  bottom=2mm,           
}
\usepackage[capitalize,noabbrev]{cleveref}

\theoremstyle{plain}
\newtheorem{theorem}{Theorem}[section]

\theoremstyle{definition}

\newtheorem{assumption}[theorem]{Assumption}
\theoremstyle{remark}

\usepackage[textsize=tiny]{todonotes}

\icmltitlerunning{Submission and Formatting Instructions for ICML 2025}

\AtBeginEnvironment{tabular}{\small}

\begin{document}

\twocolumn[
\icmltitle{No Query, No Access, Yet Dangerous: Victim Data-based Adversarial Attack}





\begin{icmlauthorlist}
\icmlauthor{Wenqiang Wang}{sysu}
\icmlauthor{Siyuan Liang}{sysu}
\icmlauthor{Yangshijie Zhang}{lzu}
\icmlauthor{Xiaojun Jia}{sysu}
\icmlauthor{Hao Lin}{sysu}
\icmlauthor{Xiaochun Cao}{sysu}
\end{icmlauthorlist}

\icmlaffiliation{sysu}{Sun Yat-sen University, Guangzhou, China}
\icmlaffiliation{lzu}{Lanzhou University, Gansu, China}


\icmlcorrespondingauthor{Xiaochun Cao}

\icmlkeywords{Machine Learning, ICML}

\vskip 0.3in
]



\printAffiliationsAndNotice{\icmlEqualContribution} 

\begin{abstract}
Textual adversarial attacks mislead NLP models, including Large Language Models (LLMs), by subtly modifying text. While effective, existing attacks often require knowledge of the victim model, extensive queries, or access to training data, limiting real-world feasibility. To overcome these constraints, we introduce the \textbf{Victim Data-based Adversarial Attack (VDBA)}, which operates using only victim texts. To prevent access to the victim model, we create a shadow dataset with publicly available pre-trained models and clustering methods as a foundation for developing substitute models. To address the low attack success rate (ASR) due to insufficient information feedback, we propose the hierarchical substitution model design, generating substitute models 
to mitigate the failure of a single substitute model at the decision boundary.
 Concurrently, we use diverse adversarial example generation, employing various attack methods to generate and select the 
adversarial example with better similarity and attack effectiveness.
Experiments on the Emotion and SST5 datasets show that VDBA outperforms state-of-the-art methods, achieving an ASR improvement of 52.08\% while significantly reducing attack queries to 0. More importantly, we discover that VDBA poses a significant threat to LLMs such as  Qwen2 and the GPT family, and achieves the highest ASR of 45.99\% even without access to the API, confirming that advanced NLP models still face serious security risks. Our codes can be found at \href{https://anonymous.4open.science/r/VDBA-Victim-Data-based-Adversarial-Attack-36EC/}{Anonymous Github}.
\end{abstract}


\begin{figure}[t]
  \centering
  \includegraphics[width=0.46\textwidth]{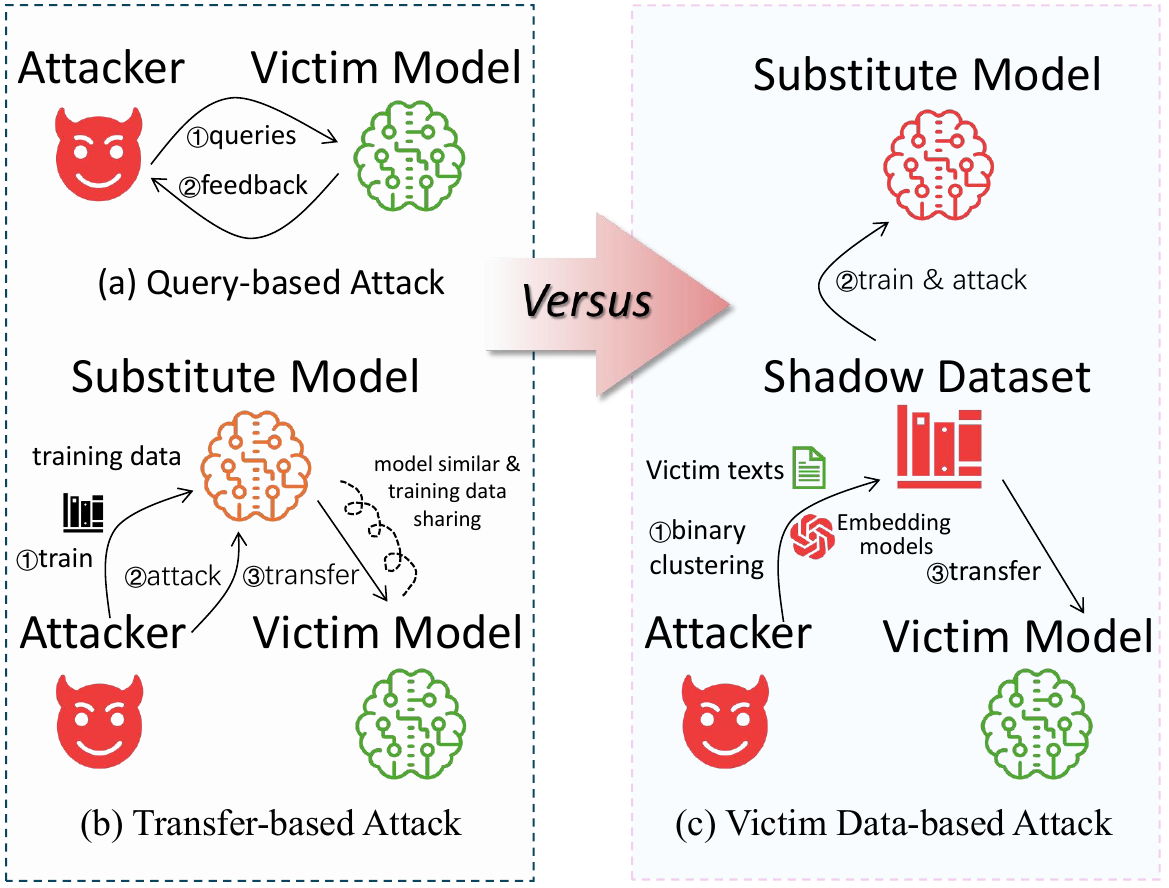}
  \vspace{-0.2in} 
  \caption{\textbf{Comparison of victim data-based, query-based, and transfer-based attacks.}
Query-based attacks generate adversarial examples by querying the victim model for feedback. Transfer-based attacks use substitute models trained on the victim's dataset to craft transferable adversarial examples. Victim data-based attacks cluster unlabeled victim texts to create a shadow dataset, training a binary classifier to generate adversarial examples.} 

  \label{Overview-of-Scenario}
  \vspace{-0.2in} 
\end{figure}

\section{Introduction}
\label{sec:intro}

Text classification, a key task in natural language processing, assigns text to categories~\cite{fields2024survey}. Neural networks, especially large language models (LLMs), have greatly improved performance~\cite{sun2023text}. However, adversarial attacks can subtly alter text, causing drastic output changes and undermining system security, with potential financial and legal consequences~\cite{wang2023punctuation, han2024bfs2adv, fursov2021adversarial, wei2018transferable,liang2020efficient, liang2022parallel, liang2022large, ying2024jailbreak, jing2025cogmorph, liang2023badclip, liang2024revisiting, liang2022imitated, chen2024less}.

Textual adversarial attacks can be categorized into query-based attacks and transfer-based attacks ~\cite{baniecki2024adversarial}. 
In query-based attacks, adversarial examples are generated by extensively querying the victim model to gather information on gradient~\cite{lin2021using} and output probability~\cite{hu2024fasttextdodger,liu2023sspattack}. variations. In contrast, transfer-based attacks involve querying a substitute model, whose decision boundary closely mirrors that of the victim model, such as a model trained on the victim model's training data. Adversarial examples generated on the substitute model can also successfully attack the victim model due to the similarity in decision boundaries between the two models, termed as the transferability of adversarial examples~\cite{kwon2022ensemble}.

However, extensive querying requires substantial time and resources, and in some real-world scenarios, the number of queries is limited. For example, the ChatGPT O1-mini~\cite{mohamad2024openai} model imposes a daily message limit of 50. Moreover, modern NLP models are typically deployed on web platforms via APIs~\cite{gonzalez2023improving}, preventing attackers from directly accessing internal model data, such as gradients and probabilities. These restrictions on query frequency and access to internal model information make precise and efficient query-based attacks impractical~\cite{liu2024hqa, hu2024fasttextdodger, wu2021improving, yuan2022adaptive, naseer2021generating, deng2021adversarial}. Additionally, the absence of a substitute model and training data further limits the effectiveness of transfer-based attacks.
In contrast to the challenges posed by extensive querying, internal model information, ready-trained substitute models, and training data, victim texts—accessible during the attack—and their attributes, observable from a single victim text, are easier to obtain. Once the victim text's attributes are identified, we can collect attribute-related texts from the internet. 
To address the challenges posed by query-based and transfer-based attacks, we propose a novel attack research question as shown in Figure \ref{Overview-of-Scenario}:  

\textit{Can textual adversarial attacks still be effectively realized in the acknowledge of \textbf{only the victim texts, including many-shot, few-shot, one-shot victim texts}}?

To achieve \textbf{zero access to the victim model}, we utilize publicly available pre-trained models to transform the victim texts into vector representations and generate binary pseudo-labels through binary clustering to construct shadow datasets to train binary substitute models. However, due to the incomplete accuracy of the labeling of the shadow dataset, the substitute model obtained from the training is not sufficient to generate effective adversarial examples. To enhance the \textit{effectiveness of the attack}, we focus on balancing attack success rate (ASR) and similarly. The Hierarchical Substitute Model Design to address
the issue of the failure of a single substitute model in attacking the victim texts at its decision boundary.
 Additionally, our Diverse Adversarial Example Generation strategy utilizes multiple attack methods to generate and select the adversarial examples with better similarity and
attack effectiveness, which maintains high attack success and similarly.

In the text classification task, VDBA can easily achieve an ASR of more than 40\%  at zero query cost. More importantly, VDBA not only helps existing attack methods to improve the ASR by 34.99\% on average without additional queries, but also poses a far-reaching threat to mainstream LLMs, such as  Qwen2, and the ChatGPT family, which can achieve up to 45.99\%  ASR. In addition, we deeply explore the attack efficacy of VDBA with very few victimized samples or severe inconsistency with the training data distribution and show that even under these extreme conditions, VDBA maintains an ASR more than 27.35\%. This demonstrates the highly destructive and indiscriminate nature of the attack against LLMs in real scenarios. Meanwhile, VDBA can achieve significant attack results even when only a small number of victim texts (\textit{e.g.}, $5$ texts) or merely the victim text attributes are available.
Our main contributions can be summarized as follows:
\ding{182} We propose a novel attack scenario for text classification tasks, revealing the possibility that an attacker can exploit potential vulnerabilities to carry out an attack without accessing the model and without knowledge of its structure and training data. 
\ding{183} We design the Victim Data-based Attack (VDBA) framework, which successfully achieves an efficient attack on the model in a completely black-box environment by constructing a shadow dataset and introducing gradient diversity. 
\ding{184} Experiments show that even the current LLMs, such as  Qwen2~\cite{yang2024qwen2}, and the ChatGPT family~\cite{an2023chatgpt}, are not able to effectively defend against such attacks, highlighting the profound threat that VDBA poses to model security.

\section{Related Work}\label{related-work}
\textbf{Text White-Box Attack:}
In a white-box text attack, attackers use full knowledge of the victim model’s structure and parameters to generate adversarial examples. For example, Hotflip~\cite{ebrahimi2018Hotflip} iteratively replaces words based on their importance score, while TextBugger~\cite{ren2019generating} perturbs both characters and words using a greedy algorithm to disrupt gradients. \textit{However, full model access limits the practicality of white-box attacks in real-world scenarios, where such information is typically unavailable.}

\textbf{Text Black-Box Attack:} In a text black-box attack, attackers generate adversarial examples without access to the model's internal structure, relying solely on input-output observations~\cite{waghela2024modified,han2024bfs2adv,zhu2024limeattack,kang2024hybrid}.
The attacker identifies important words based on the model’s output probabilities and targets them sequentially to alter the label. Examples include SememePSO~\cite{zang2020word}, which optimizes the search space using metaheuristics, and BAE~\cite{garg2020Bae}, which replaces words using BERT and prompt learning. Leap~\cite{xiao2023leap} speeds up convergence with adaptive particle swarm optimization, while HQA~\cite{liu2024hqa} minimizes perturbations by selecting optimal synonyms.
\textit{However, real-world systems often limit access to model details and frequencies, presenting new challenges for black-box attacks.}
\begin{figure*}[t]
  \centering
  \includegraphics[width=0.993\textwidth]{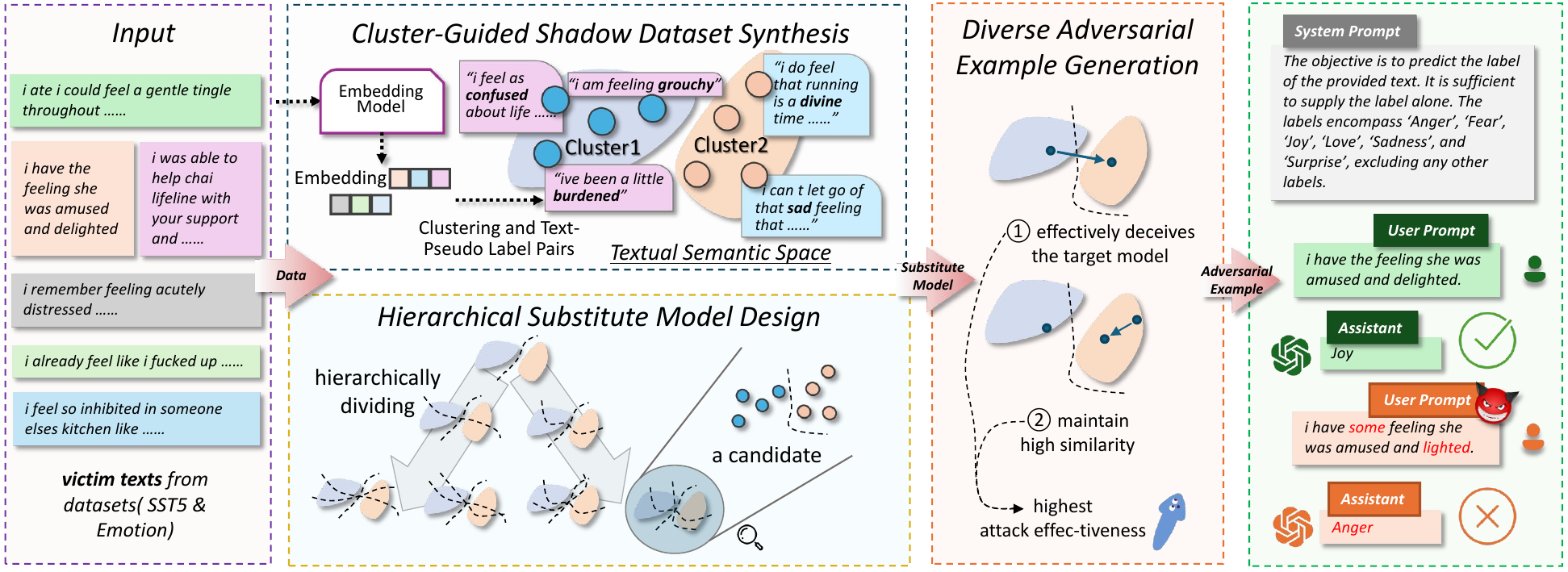}
  \vspace{-0.20in} 
  \caption{\textbf{The overview of the Victim Data-based  Attack (VDBA).}  In Cluster-Guided Shadow Dataset Synthesis, VDBA utilizes an embedding model to vectorize victim texts, followed by binary clustering to generate pseudo-labels, thus creating the shadow dataset. In Hierarchical Substitute Model Design, VDBA further clusters texts with identical labels, hierarchically training a new substitute model. This approach provides a broader range of substitute model options, thereby improving the success rate of adversarial attacks. In Diverse Adversarial Example Generation, VDBA employs multiple attack algorithms to generate candidate adversarial examples, selecting the final example based on changes in probability and similarity.
}
  \label{Overview-of-all}
  \vspace{-0.20in}
\end{figure*}

In summary, the need for internal model information makes text white-box attacks challenging to implement in real-world scenarios, while restrictions on probability scores,  and query limits may hinder the effectiveness of black-box attacks. \textbf{In contrast, VDBA is well-suited for more practical attack environments, as it can generate adversarial examples with high success rates even when only the victim texts or victim texts' topic is available.}

\section{Preliminary and Definition}

\textbf{Victim Model, Victim Texts, and Substitute Model }: The victim model refers to the target model that the attacker intends to attack, denoted as $f_{\text{v}}$. 
 The \text{victim texts} refers to the text that the attacker intends to modify. Attackers can access these texts when attacking them. 
 The substitute model approximates the decision boundary of the victim model and is used to generate adversarial samples, which the training data can train, denoted as $f_{\text{s}}$.




 \textbf{Victim Data-based Adversarial Attack:}
 Attackers cannot access the victim model $f_{\text{v}}$, and training data. Attackers craft the adversarial examples just based on the victim texts $(\mathbf{D}_{\text{v}} = \{\mathbf{x}_{1}, \mathbf{x}_{2}, \dots, \mathbf{x}_{n}\})$ without known labels. Attackers train the binary substitute model $f_{\text{s}}$ from the victim texts as the following formal definition:
\begin{equation}
\begin{aligned}
& \mathbf{D} = \{(\mathbf{x}_1, y_1^{p}), (\mathbf{x}_2, y_2^{p}), \dots, (\mathbf{x}_n, y_n^{p})\}, \\
& \mathbf{\theta^*} = \arg\min_{\mathbf{\theta}} \frac{1}{|\mathbf{D}|} \sum_{(\mathbf{x}_i, y_i^p) \in \mathbf{D}} (y_i^p - f_{\text{s}}(\mathbf{\theta}; \mathbf{x}_i))^2,\\
\end{aligned}
\end{equation}
where $y_i^p$ is the pseudo label of $\mathbf{x}_i$, $\mathbf{\theta^*}$ is the model parameters of the substitute model.
The perturbed text \( \tilde{\mathbf{x}}_i \) is generated according to the following formal definition:
\begin{equation}\label{Adv-attack}
\begin{split}
 & \tilde{\mathbf{x}}_i  = \arg\max_{\mathbf{x}_i} L(f_{\text{s}}(\tilde{\mathbf{x}}_i),  f_{\text{s}}(\mathbf{x}_i)), 
\quad  \text{and} \quad f_{\text{v}}(\mathbf{x}_i) \neq f_{\text{v}}(\tilde{\mathbf{x}}_i), \\
&\quad \text{s.t.} \quad \|\tilde{\mathbf{x}}_i - \mathbf{x}_i\|_p \leq \epsilon.
\end{split}
\end{equation}
where  \( L(f(\tilde{\mathbf{x}}_i), f(\mathbf{x}_i)) \) is the loss function, such as cross-entropy loss~\cite{mao2023cross}, \( f_{\text{s}}(\tilde{\mathbf{x}}_i) \) is the substitute model's prediction for \( \tilde{\mathbf{x}}_i \), and \( f_{\text{s}}(\mathbf{x}_i) \) is  the substitute model's prediction for $x_i$, \( \epsilon \) is the perturbation size limit. \( \|\tilde{\mathbf{x}}_i - \mathbf{x}_i\|_p \leq \epsilon \) limits the distance between the adversarial example and the original example, such as \( l_2 \)  norm.


\section{Victim Data-based Attack}
As shown in Figure \ref{Overview-of-all}, DBA comprises three components: Cluster-Guided Shadow Dataset Synthesis, which generates the shadow dataset; Hierarchical Substitute Model Design, which hierarchically trains substitute models to ensure that each victim model has a suitable substitute model and mitigates the failure of a single substitute model
at the decision boundary; and Diverse Adversarial Example Generation, which generates multiple candidate adversarial examples and selects the final one based on high similarity and high probability changes.

\subsection{Cluster-Guided Shadow Dataset Synthesis}\label{cluster-number}

\begin{figure}
  \centering
  \includegraphics[width=0.485\textwidth]{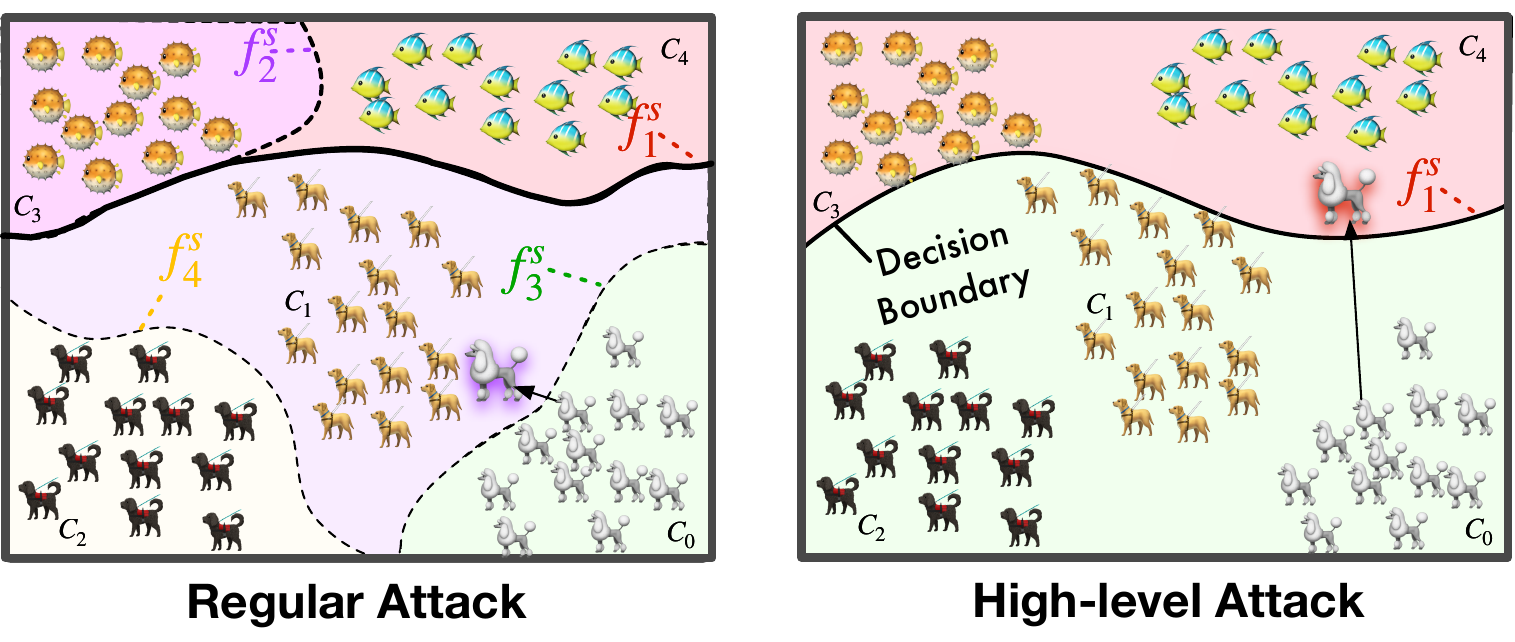} 
     \vspace{-0.35in}
  \caption{\textbf{The overview of High-level Attack.}
A Regular Attack encompasses both query-based and transfer-based attacks, aiming to misclassify a corgi as a non-corgi.
In contrast,  High-level Attack first clusters victim data and then trains a substitute model to misclassify a corgi from the ``dog'' cluster to the ``non-dog'' cluster, which can craft adversarial examples using only victim data.
  }
  \label{Overview-high-level}
  \vspace{-0.20in}
\end{figure}

Training a substitute model without access to the original training data $\mathbf{D}^{\text{tr}} = \{(\mathbf{x}_1^{\text{tr}}, y_1^{\text{tr}}), (\mathbf{x}_2^{\text{tr}}, y_2^{\text{tr}}), \dots, (\mathbf{x}_n^{\text{tr}}, y_n^{\text{tr}})\}.$
 poses a significant challenge in adversarial settings.
This section proposes a method to synthesize a shadow dataset by assigning pseudo-labels to victim texts.
Due to the inaccessibility of the victim model and the ground-truth labels of the victim texts, we opt to extract high-level labels directly from the victim texts to train the substitute model. High-level labels are more generalizable compared to the actual labels. As Figure  \ref{Overview-high-level} shows: ``dog'' is a high-level label compared to more specific labels such as ``Doberman,'' ``Bulldog,'' and ``Corgi.'' Meanwhile, we propose the high-level attack hypothesis: 

\begin{assumption}[\textbf{High-level Attack Hypothesis}]

The adversarial example generated in the 
 substitute model $f_{\text{s}}$ trained with high-level labels can also successfully attack the victim model $f_{\text{v}}$.
 Formally,
 \begin{equation}
 f_{\text{s}}(\mathbf{x}) \neq f_{\text{s}}(\tilde{\mathbf{x}}),
\text{and} \quad f_{\text{v}}(\mathbf{x}) \neq f_{\text{v}}(\tilde{\mathbf{x}}), 
\end{equation}
\end{assumption}
where the $\tilde{\mathbf{x}}$ is the adversarial example of $\mathbf{x}$.
For example, if an image labeled as ``Corgi'' in the victim model is used, and the adversarial example generated by the substitute model \( f_{\text{s}} \)—which is trained with high-level labels—causes a shift in the label from ``Dog'' to a ``Non-Dog'' category, the resulting adversarial example will also be farther from the ``Corgi'' label in the victim model. Therefore, even though \( f_{\text{s}} \) and \( f_{\text{v}} \) may be dissimilar, adversarial examples generated using high-level labels in \( f_{\text{s}} \) can still successfully deceive the victim model \( f_{\text{v}} \).




To utilize victim texts effectively, we apply the clustering method K-means~\cite{lloyd1982least} to assign pseudo-labels to these unlabeled texts, thus creating a shadow dataset. To obtain the highest-level labels, we chose to set the number of clusters to 2, as our training goal is to distinguish between class and non-class. We examine the impact of the number of clusters on the substitute model in Section \ref{Ablation_Study}. 
As the Figure \ref{Overview-of-all} shows,
 each victim text \(\mathbf{x}_i\) is embedded using an embedding method, such as the pre-trained model or one-hot vectors, \(f_{\text{e}}\), denoted as \(\mathbf{e}_i = f_{\text{e}}(\mathbf{x}_i)\). All victim text embeddings \(\mathbf{E} = \{\mathbf{e}_1, \mathbf{e}_2, \dots, \mathbf{e}_n\}\) are then clustered into two distinct groups, and cluster labels $y_i^{p}$ are assigned as pseudo-labels for each text. The resulting shadow dataset \(\mathbf{D}\) is defined as:
 \begin{equation}
\mathbf{D} = \{(\mathbf{x}_1, y_1^{p}), (\mathbf{x}_2, y_2^{p}), \dots, (\mathbf{x}_n, y_n^{p})\}. 
\end{equation}
Despite the discrepancies between victim data-pseudo label pairs and the original training data $\mathbf{D}^{\text{tr}}$, which can hinder the perfect alignment of the substitute model with the victim model's decision boundary, our aim is not to mirror the victim model but to develop a substitute model that can distinguish between classes clearly for high-level attack. The substitute model, trained on the shadow dataset \(\mathbf{D}\), while not a replicate of the victim model, is adequate for conducting effective adversarial attacks.

\subsection{Hierarchical Substitute Model Design}

The initial substitute model \( f_1^s \), trained on the victim dataset \(\mathbf{D}\), often struggles to attack victim texts near its decision boundary. For instance, texts with a \textit{neutral} ground truth label are frequently clustered into broader sentiment groups such as positive (\textit{happy}, \textit{surprise}) or negative (\textit{anger}, \textit{fear}). This results in coarse decision boundaries, making \( f_1^s \) less effective. Even when adversarial perturbations alter the clustering label (e.g., from \( C_{\text{pos}} \) to \( C_{\text{neg}} \)), the victim model \( f_v \) might still predict the original label (\textit{neutral}), causing the attack to fail. This suggests that the current substitute model is not a suitable substitute model for victim texts situated near its decision boundary.

\begin{equation}
    \tilde{P}_M(\mathbf{y}_t \mid \mathcal{S}_t, \hat{\mathbf{x}}) = \frac{P_M(\mathbf{y}_t \mid \mathcal{S}_t, \hat{\mathbf{x}})}{P_M(\mathbf{y}_{gt} \mid \mathcal{S}_t, \hat{\mathbf{x}})+P_M(\mathbf{y}_t \mid \mathcal{S}_t, \hat{\mathbf{x}})}.
\end{equation}

Moreover, failed attacks are often associated with minimal changes in the predicted probabilities of the original and adversarial texts. For example, if the predicted probability of the original text for its label is 0.52 and drops to 0.47 after the perturbation, the small difference (0.05) leads to an ineffective attack. To overcome these challenges, we propose a hierarchical approach to improve substitute model performance and enhance attack success rates.

\begin{figure}
  \centering
\includegraphics[width=0.485\textwidth]{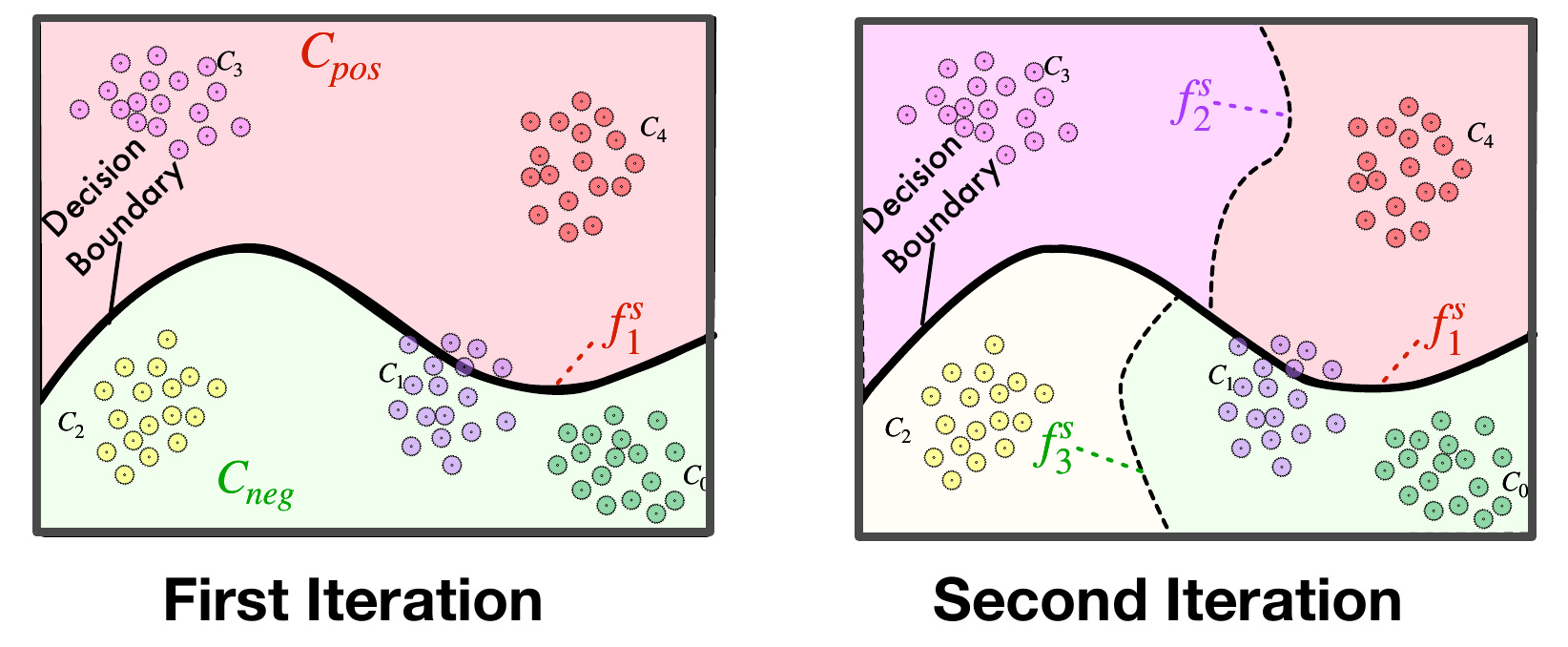} 
     \vspace{-0.35in}
  \caption{\textbf{The overview of Hierarchical Substitution Model Design.} For the first iteration of VDBA, all victim texts are employed to 
 train the substitute model $f_1^s$. For the second iteration, the victim texts with the positive cluster label $C_{\text{pos}}$ are employed to 
 train the substitute model $f_2^s$, the victim texts with the negative cluster label $C_{\text{neg}}$ are employed to 
 train the substitute model $f_3^s$. Following this process, we can train the hierarchical substitution models.}
  \label{Overview-of-Hierarchical}
  \vspace{-0.20in}
\end{figure}


To address these limitations, we propose a hierarchical substitute model design. The process begins by training \( f_1^{\text{s}} \) on the victim dataset. As Figure \ref{Overview-of-Hierarchical} shows, texts with cluster label $C_{\text{pos}}$ and $C_{\text{neg}}$ are then re-clustered into finer-grained cluster categories, enabling the training of additional substitute models, such as \( f_2^\text{s} \) and \( f_3^\text{s} \). These models offer victim texts a broader range of substitute model options, enabling the selection of a suitable substitute model to enhance the success rate of adversarial attacks.
Iteratively refining clusters and training new substitute models allows for incremental improvements in the adversarial effectiveness. Meanwhile, $u$ attack iteration will generate $U$ substitute model, where $U=2^0 + 2^1 + 2^2 + \dots + 2^u$.

We use Theorem~\ref{max_prob1} to formalize the intuition that as the number of substitute models increases, the probability that at least one adversarial example generated by these models successfully attacks the victim model also increases:

\begin{theorem}\label{max_prob1}
Let \( \{p_1^s, p_2^s, \dots, p_m^s\} \) represent the success probabilities of attacking the target model, with the cumulative distribution function (CDF) denoted as \( F(p) \),  The complementary probability is $\Pr(p_i^s > p) = 1 - F(p)$. As \( m \to \infty \), the probability that \( \max\{p_1^s, p_2^s, \dots, p_m^s\} > p_i^s \) approaches 1 for any \( p_i^s < 1 \). Formally:
\begin{equation}
\small
\begin{array}{c}
\lim\limits_{m \to \infty} \Pr\left( \max\{ p_1^s, p_2^s, \dots, p_m^s \} > p_i^s \right) = 1, \\
\lim\limits_{m \to \infty} \Pr\left( \max\{ p_1^s, p_2^s, \dots, p_m^s \} > \max\{ p_1^s, p_2^s, \dots, p_{m_1}^s \} \right) = 1, \\
\text{for} \quad m > m_1.
\end{array}
\end{equation}
\end{theorem}


\textit{Proof.} Please refer to Appendix~\ref{proof_max_prob1} for details.

This hierarchical approach enables the creation of multiple substitute models to mitigate the failure of a single substitute model
at the decision boundary. By leveraging these models, the proposed framework ensures higher attack success rates, even for victim texts that are challenging to perturb effectively.

     .

\subsection{Diverse Adversarial Example Generation}\label{Diverse}

\begin{figure}
  \centering
  \includegraphics[width=0.485\textwidth]{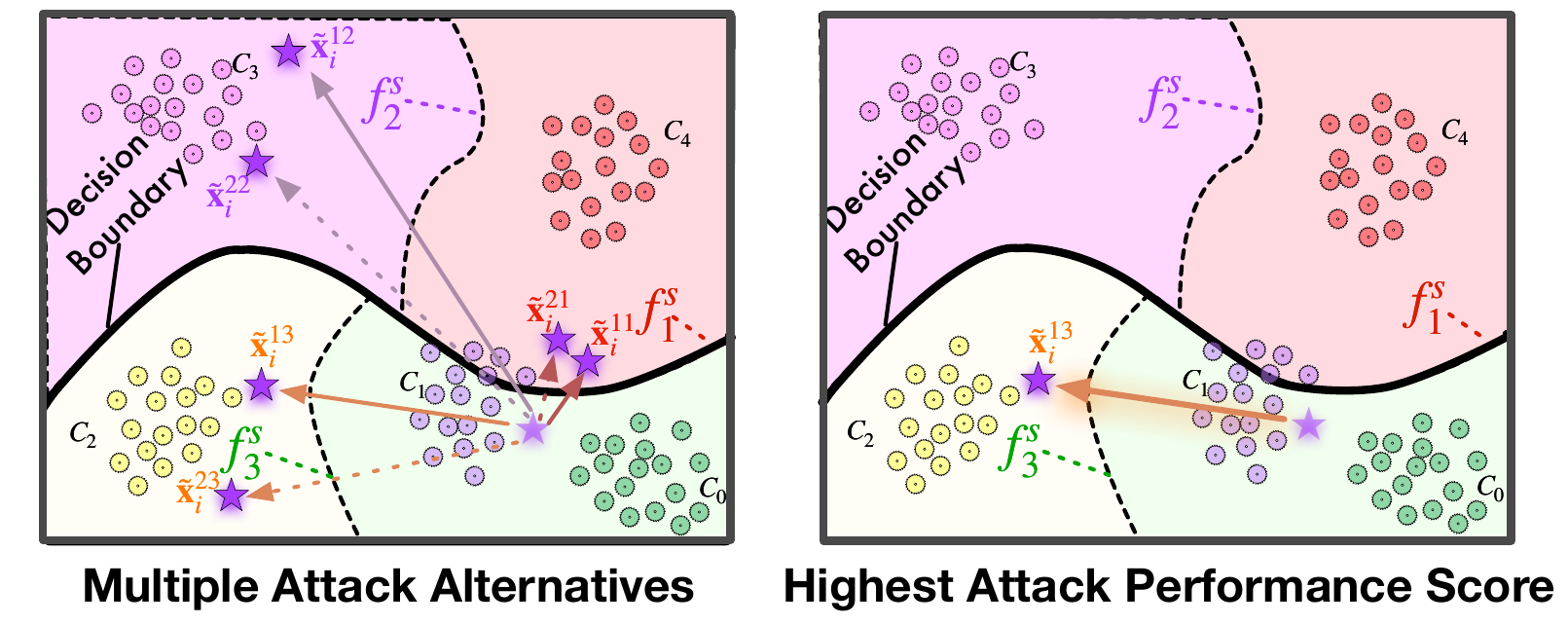} 
     \vspace{-0.35in}
  \caption{\textbf{The overview of Diverse Adversarial Example Generation.} 
The left figure depicts the generation of six candidate adversarial samples using two attack methods ($M1$, $M2$) and three substitute models ($f_1^{\text{s}}$, $f_2^{\text{s}}$,$f_3^{\text{s}}$).
According to Equation \ref{select_adv}, $\tilde{\mathbf{x}}_i^{11}$ and $\tilde{\mathbf{x}}_i^{21}$ are eliminated for violating the constraint:$
p_{\hat{y}}^{f_j^{\text{s}}}(\mathbf{x}_i^{lj}) - p_{\hat{y}}^{f_j^{\text{s}}}(\tilde{\mathbf{x}}_i^{lj}) \geq \epsilon^{'}
$ which suggests that $\mathbf{x}_i$ is close to the decision boundary of $f_1^{s}$, rendering it an unsuitable substitute model.
Similarly, $\tilde{\mathbf{x}}_i^{12}$ and $\tilde{\mathbf{x}}_i^{22}$ are excluded due to insufficient similarity, failing to satisfy:
$
\frac{f_{\text{e}}(\mathbf{x}) \cdot f_{\text{e}}(\tilde{\mathbf{x}}_i^{lj})}{\|f_{\text{e}}(\mathbf{x})\| \cdot \|f_{\text{e}}(\tilde{\mathbf{x}}_i^{lj})\|}  \geq \epsilon^{*}
$
The remaining candidates, $\tilde{\mathbf{x}}_i^{13}$ and $\tilde{\mathbf{x}}_i^{23}$, are assessed according to their attack performance scores $s_{ilj}$. The sample with the highest score, $\tilde{\mathbf{x}}_i^{13}$, is chosen as the final adversarial example for $\mathbf{x}_i$. 
  }
  \label{Overview-of-Diverse-Adversarial-Example}
  \vspace{-0.15in}
\end{figure}

This section explores how to select optimal attack strategies for each victim text with its suitable substitute models.

Given a victim text $\mathbf{x}_i$, our goal is to generate an adversarial example $\tilde{\mathbf{x}}_i$ that effectively deceives the target model while maintaining high similarity. Instead of relying on a single attack method, we explore leveraging multiple attack methods and substitute models to maximize attack effectiveness.

Let $\{M_1, M_2, \dots, M_w\}$ be $w$ different adversarial attack methods, and $\{f_1^{\text{s}}, f_2^{\text{s}}, \dots, f_U^{\text{s}}\}$ be $U$ substitute models. By applying each attack method to each substitute model, we generate a total of $wU$ candidate adversarial examples:
\begin{equation}
    \{\tilde{\mathbf{x}}_i^{11}, \tilde{\mathbf{x}}_i^{21}, \dots, \tilde{\mathbf{x}}_i^{w1}, \dots, \tilde{\mathbf{x}}_i^{wU}\}, \quad \tilde{\mathbf{x}}_i^{lj} = M_l(\mathbf{x}_i, f_j^{\text{s}}).
\end{equation}

To formalize the advantage of using multiple attacks, let $P^{\text{sim}}_{(w)}$ denote the probability that at least one generated adversarial example surpasses a predefined similarity threshold $\epsilon^{*}$. Theorem \ref{max_prob2} states:
\begin{theorem}\label{max_prob2}
Suppose $\{ p_1^{\text{sim}}, p_2^{\text{sim}}, \dots, p_w^{\text{sim}} \}$ represent the probabilities that adversarial examples from methods $\{ M_1, M_2, \dots, M_w \}$ satisfy the similarity constraint. Let $\mu_w$ denote the expected number of valid adversarial examples. Then:
\begin{equation}
    \lim_{w \to \infty} P^{\text{sim}}_{(w)} = 1, \quad \mu_w > \mu_h, \quad P^{\text{sim}}_{(w)} > P^{\text{sim}}_{(h)}, \quad \text{for} \quad w > h.
\end{equation}
\end{theorem}
\textit{Proof.} See Appendix \ref{proof_max_prob2}.

This result indicates that increasing the number of attack methods enhances the probability of generating high-quality adversarial examples. Among the $wU$ candidate adversarial examples, we select the one with the highest attack effectiveness, defined by:
\begin{equation}\label{select_adv}
\small
\begin{split}
 & s_{ilj} =  \alpha \left( p_{\hat{y}}^{f_j^{\text{s}}}(\mathbf{x}_i^{lj}) - p_{\hat{y}}^{f_j^{\text{s}}}(\tilde{\mathbf{x}}_i^{lj}) \right) + \beta \left( \frac{f_{\text{e}}(\mathbf{x}_i) \cdot f_{\text{e}}(\tilde{\mathbf{x}}_i^{lj})}{\|f_{\text{e}}(\mathbf{x}_i)\| \cdot \|f_{\text{e}}(\tilde{\mathbf{x}}_i^{lj})\|} \right), \\
& \text{s.t.} \quad 
    p_{\hat{y}}^{f_j^{\text{s}}}(\mathbf{x}_i^{lj}) - p_{\hat{y}}^{f_j^{\text{s}}}(\tilde{\mathbf{x}}_i^{lj}) \geq \epsilon^{'},
    \quad \frac{f_{\text{e}}(\mathbf{x}) \cdot f_{\text{e}}(\tilde{\mathbf{x}}_i^{lj})}{\|f_{\text{e}}(\mathbf{x})\| \cdot \|f_{\text{e}}(\tilde{\mathbf{x}}_i^{lj})\|}  \geq \epsilon^{*},
\end{split}
\end{equation}
where $p_{\hat{y}}^{f_j^{\text{s}}}(\mathbf{x}_i^{lj}) - p_{\hat{y}}^{f_j^{\text{s}}}(\tilde{\mathbf{x}}_i^{lj})$ quantifies the impact of the attack on the predicted probability of the original class $\hat{y}$ with the substitute model $f_j^{\text{s}}$, high predicted probability change ($\geq \epsilon^{'}$) avoid locating the  decision boundary of the substitute model. The term $\frac{f_{\text{e}}(\mathbf{x}_i) \cdot f_{\text{e}}(\tilde{\mathbf{x}}_i^{lj})}{\|f_{\text{e}}(\mathbf{x}_i)\| \cdot \|f_{\text{e}}(\tilde{\mathbf{x}}_i^{lj})\|}$ measures the semantic similarity between $\mathbf{x}_i$ and $\tilde{\mathbf{x}}_i^{lj}$, high similarity ($\geq \epsilon^{*}$) ensures that the adversarial example retains the same true label as the original text. The parameters $\alpha$ and $\beta$ control the balance between attack strength and similarity constraints. $\epsilon^{'}$ and $\epsilon^{*}$ are the thresholds.
The final adversarial example $\tilde{\mathbf{x}}_i$ is the one maximizing $s_{ilj}$ under the given constraints, ensuring both attack effectiveness and similarity.
We present a detailed adversarial example selection process in Figure \ref{Overview-of-Diverse-Adversarial-Example}.

\section{Experiment}

\begin{table*}[t]

\centering
\caption{The attack performance of VDBA and adversarial attacks on Emotion and SST5 datasets with many-shot victim texts.  For each metric, the best method is highlighted in \textbf{bold} and the runner-up is \underline{underlined}. }
\label{main-tab}
\resizebox{0.995\textwidth}{!}{

\begin{tabular}{@{}ccccccc|cccccc@{}}
\toprule
\multirow{3}{*}{Method} & \multicolumn{6}{c|}{SST5}                                        & \multicolumn{6}{c}{Emotion}                                     \\ \cmidrule(l){2-13} 
                        & \multicolumn{3}{c}{DistilBERT}    & \multicolumn{3}{c|}{RoBERTa} & \multicolumn{3}{c}{DistilBERT}    & \multicolumn{3}{c}{RoBERTa} \\ \cmidrule(l){2-13} 
                        & ASR(\%) $\uparrow$ & Sim $\uparrow$ & Queries $\downarrow$               & ASR(\%) $\uparrow$ & Sim $\uparrow$ & Queries $\downarrow$               & ASR(\%) $\uparrow$ & Sim $\uparrow$ & Queries $\downarrow$               & ASR(\%) $\uparrow$ & Sim $\uparrow$ & Queries $\downarrow$               \\ \midrule \midrule
 Bae& 42.71&     0.888 
& \multicolumn{1}{c|}{47360
} &        39.14   
&        0.887   
&            47466  
& 32.25 
&     0.926 
& \multicolumn{1}{c|}{43682
} &        32.95   
&        0.923  
&           43656  
\\
FD                      & 25.20 
&     0.939 
& \multicolumn{1}{c|}{27760
} &        22.30   
&        \textbf{0.982}&            21452  &

22.30 
&     0.932 
& \multicolumn{1}{c|}{25612
} &        27.50    
&        \textbf{0.982}&           22850  
\\
 Hotflip& 41.50&     0.951 
& \multicolumn{1}{c|}{25455
} &        29.00   
&        0.951   
&            25956  
& 29.00 
&     0.949 
& \multicolumn{1}{c|}{28566
} &        28.05   
&        0.949  
&           28800  
\\
PSO                     & 45.14 
&     0.954 
& \multicolumn{1}{c|}{24398
} &        \underline{41.50}&        0.954   
&            27360  &

39.50 
&     \underline{0.952}& \multicolumn{1}{c|}{23660
} &        \underline{37.65}&        0.951  
&           24190  
\\
 TextBug                 & 30.36&     \textbf{0.978}& \multicolumn{1}{c|}{69520
} &        20.85   
&        \underline{0.978}&            67009  
& 20.85 
&     \textbf{0.978}& \multicolumn{1}{c|}{60642
} &        21.45   
&        \underline{0.978}&           60662  \\
Leap                    & 32.55      
&     \underline{0.953}& \multicolumn{1}{c|}{\underline{21548}} &        30.07      
&        0.944               
&            \underline{21083}&

40.58      
&     0.926               
& \multicolumn{1}{c|}{\underline{19460}} &        37.63      
&        0.911               
&           \underline{}19560\\
 CT-GAT                    & 
29.37&     0.939               
& \multicolumn{1}{c|}{46238            
} &        24.80      
&        0.926               
&            82957            
& 28.10      
&     0.904               
& \multicolumn{1}{c|}{52114            
} &        30.85      
&        0.906               
&           50686            
\\
 HQA                     & \underline{46.11}&     0.936               
& \multicolumn{1}{c|}{64855            
} &        39.64      
&        0.929               
&            64256            
&

\underline{37.40}&     0.912               
& \multicolumn{1}{c|}{44876            
} &        
36.40      
&        0.911               
&           46326            
\\
\midrule \rowcolor{orange!50}  \textbf{VDBA}                     & 
\textbf{52.08}&   0.950& \multicolumn{1}{c|}{\textbf{0}} &        \textbf{45.03}&       0.950&            \textbf{0}& \textbf{43.15}&     0.949& \multicolumn{1}{c|}{\textbf{0}} &       \textbf{42.05}&        0.949&          \textbf{0}\\ \bottomrule
\end{tabular}
}
\end{table*}
\subsection{Experimental Setup}

\textbf{Datasets, metrics, and substitute model.} (1) We perform primary experiments on the SST5~\cite{socher2013recursive} and Emotion~\cite{saravia2018carer} datasets. The details of the datasets are presented in Appendix \ref{datasets}. 
(2) Several metrics are used to evaluate the effectiveness of VDBA, including Attack Success Rate (ASR), Semantic Similarity (Sim), and Queries. 
``Sim'' represents the similarity between the original texts and adversarial examples, with higher ``Sim'' values indicating better attack results. ``Queries'' denotes the total number of queries made by the attack method to the model in attacking time, where fewer ``Queries'' correspond to lower query costs and improved attack performance.
Detailed descriptions of these evaluation metrics are provided in Appendix \ref{em}.
(3)
The details of the substitute model training and  architecture are presented in Appendix \ref{surrogate-model}.

\textbf{Baselines, clustering and embedding method.}
In the victim texts-only scenario, no prior work on attacks has been carried out. 
Therefore,
we have broadened the application conditions of other methods, enabling other text-attack algorithms to access any required information, including losses, gradients, and Probabilities. Several text attack methods are selected, including Bae~\cite{garg2020Bae},  FD~\cite{papernot2016crafting}, Hotflip~\cite{ebrahimi2018Hotflip}, PSO~\cite{zang2020word}, TextBug~\cite{ren2019generating}, Leap~\cite{xiao2023leap}, CT-GAT~\cite{lv2023ct}, and HQA~\cite{liu2024hqa}. The details of these methods are presented in Appendix \ref{baseline}.  To ensure a fairer comparison, we restrict other attack methods to a maximum of $35$ accesses per victim text, as VDBA does not access the victim model at all.
We utilize K-means~\cite{lloyd1982least} as the clustering method and the pre-trained embedding model ``T5''~\cite{raffel2020exploring} as the embedding method.

\textbf{Attack methods and 
hyperparameter in VDBA.}
(1) 
VDBA employs BAE~\cite{garg2020Bae}, FD~\cite{papernot2016crafting}, Hotflip~\cite{ebrahimi2018Hotflip}, PSO~\cite{zang2020word}, and TextBug~\cite{ren2019generating} as the attack methods. Furthermore, the configurations of these methods are identical to those used in the Baselines.
(2)
VDBA conducts $3$ iterations of the hierarchical substitute model training and trains $8$ substitute models.
The scaling factors \( \alpha \) and \( \beta \) in Equation \eqref{select_adv} are set to $3$ and $1$, respectively. Similarly, the thresholds \( \epsilon^{'} \) and \( \epsilon^{*} \) in Equation \eqref{select_adv} are set to $0.35$ and $0.85$, respectively.



\subsection{Attack Results with  Victim Texts}\label{com_querry}
We evaluate VDBA under three scenarios: Many-shot, Few-shot, and One-shot victim texts. In the one-shot scenario, clustering is infeasible. Instead, we analyze the attributes of the single victim text, retrieve related texts from online sources, and use them to train substitute models. This scenario highlights a dataset domain-shift problem. We further investigate how the number of victim texts and the use of texts from different domains impact VDBA performance.
In the \textbf{Many-shot} scenario, attackers have full access to all victim texts, including 2,210 from SST5 and 2,000 from Emotion.  
In the \textbf{Few-shot} scenario, access is limited to subsets of 5, 10, 20, or 40 victim texts from SST5 and Emotion.  
In the \textbf{One-shot} scenario, given that Emotion and SST5 are 5-class and 6-class sentiment datasets, respectively, we utilize the 28-class Go-Emotions dataset~\cite{Goemotion} and the 2-class Rotten Tomatoes dataset~\cite{Pang+Lee:05a} as sources of related texts.



\textbf{Many-shot victim texts.} 
Table \ref{main-tab} presents experimental results for the SST5 and Emotion datasets with many-shot victim texts. VDBA achieves state-of-the-art (SOTA) performance in both ASR and Queries metrics, demonstrating superior attack success rates with zero queries during attack. Notably, VDBA achieves these results without accessing the training data or querying the victim model. While semantic similarity may be slightly reduced during the hierarchical substitution model process, our diverse adversarial example generation method ensures a relatively high level of semantic similarity. Other attack methods can improve ASR by increasing query costs, but this comes with a significant query overhead, making them impractical for real-world use. Results without query limitations are presented in Appendix \ref{unlimited}. To evaluate VDBA's applicability to other text classification tasks, we tested it on AG News~\cite{Zhang2015CharacterlevelCN} for news classification and TREC6~\cite{voorhees2000overview} for question classification. As shown in Table \ref{More-data-results} (Appendix \ref{more-result}), VDBA achieves SOTA results on both datasets.

\begin{table}[t]
\centering
\caption{Few-shot attack performance of VDBA on Emotion and SST5 datasets. VDBA can achieve significant ASR even in scenarios with very few victim texts. }
\label{few-shot}
\resizebox{0.48\textwidth}{!}{%
\begin{tabular}{@{}c|ccc|ccc@{}}
\toprule
          & \multicolumn{3}{c|}{Emotion}                                                                                                        & \multicolumn{3}{c}{SST5}                                                                                                            \\ \midrule
Shot Size & \begin{tabular}[c]{@{}c@{}}Roberta\\ ASR(\%) ↑\end{tabular} & \begin{tabular}[c]{@{}c@{}}Distilbert\\ ASR(\%) ↑\end{tabular} & Sim ↑ & \begin{tabular}[c]{@{}c@{}}Roberta\\ ASR(\%) ↑\end{tabular} & \begin{tabular}[c]{@{}c@{}}Distilbert\\ ASR(\%) ↑\end{tabular} & Sim ↑ \\ \midrule  \midrule
5-shot    &                                                             27.55   &                                                                27.35   &      0.929   &                                                             36.53   &                                                                45.27   &      0.903   \\
10-shot    &                                                             29.75   &                                                                30.00   &      0.923   &                                                             37.66   &                                                                46.36   &      0.906   \\
20-shot    &                                                             30.15   &                                                                29.30   &      0.923   &                                                             37.51   &                                                                46.06   &      0.902   \\
40-shot    &                                                             33.25   &                                                                33.30   &      0.916   &                                                             38.51   &                                                                46.61   &      0.905   \\
Full-shot &                                                             42.05&                                                                43.15&      0.949&                                                             45.03   &                                                                52.08   &      0.950       \\ \bottomrule
\end{tabular}

\vspace{-25pt}

}
\end{table}

\textbf{Few-shot victim texts.}
The attack results for the few-shot data are presented in Table \ref{few-shot}. Even with as few as five victim texts, the VDBA algorithm achieves ASR values of 27.55\%, 27.35\%, 36.53\%, and 45.27\%, respectively. Notably, when the sample size is extremely small, fewer samples can occasionally yield higher ASR, suggesting that ASR fluctuates significantly with very limited data. Furthermore, reducing the number of victim texts generally lowers similarity, but this reduction also exhibits fluctuations under scenarios with extremely few victim texts. These findings demonstrate that VDBA remains effective in attacking the victim model, even when the amount of victim text is highly constrained.

\textbf{One-shot victim text.}
 The results in Table \ref{Zero-shot-Data} demonstrate that even with only one-shot victim text, VDBA achieves an ASR exceeding 30\%. For example, on the SST5 dataset, when DistilBERT is used as the victim model and Go-Emotions data are sourced online, the ASR reaches 50.36\%. 
Comparable to the 52.08\% achieved using all SST5 victim texts directly, 50.36\% is a remarkably good result with accessing one-shot victim text.
While domain-shifted data collected online reduce the ASR, it remains significantly high. Notably, the one-shot scenario has a more pronounced negative impact on similarity compared to ASR. These findings highlight that VDBA can effectively generate adversarial examples for victim texts using domain-shifted data, provided the collected data share similar textual attributes.

\begin{table}[t]
\centering
\caption{One-shot attack performance of VDBA.
VDBA can achieve significant ASR even without accessing the victim texts.}

\label{Zero-shot-Data}
\resizebox{0.48\textwidth}{!}{%
\begin{tabular}{@{}ccccc@{}}
\toprule
Victim   Dataset    & Access Dataset & \begin{tabular}[c]{@{}c@{}}Roberta\\ ASR(\%) ↑\end{tabular} & \begin{tabular}[c]{@{}c@{}}Distilbert\\ ASR(\%) ↑\end{tabular} & Sim↑ \\ \midrule \midrule
\multirow{3}{*}{SST5}& Go-emotion & 39.46     & 50.36     & 0.911     \\
                   & Tomatoes  & 40.00   & 49.82   & 0.909   \\
                   & SST5& 45.03   & 52.08   & 0.950   \\ \midrule
\multirow{3}{*}{Emotion}& Go-emotion 
& 30.15     & 33.60     & 0.932     \\
                   & Tomatoes  
& 31.05   & 32.45   & 0.927   \\
                   & Emotion & 42.05   & 43.15   & 0.949   \\ \bottomrule
\end{tabular}
}
\vspace{-25pt}
\end{table}
\textbf{Discussion.} VDBA shows strong attack performance across many-shot, few-shot, and one-shot victim text scenarios. While effectiveness decreases with fewer victim texts, it remains notable. With very limited victim texts, performance fluctuates based on availability. In one-shot scenarios, training substitute models with online attribute-related data causes a domain shift, slightly reducing attack performance, though less so than in the few-shot case.

\subsection{Ablation Study }\label{Ablation_Study}

To explore factors influencing VDBA attack performance, including cluster numbers, training iterations of the hierarchical substitute model, attack methods, and clustering and embedding techniques, we conduct an ablation study. We evaluated the impact of cluster numbers by testing 2, 3, and 4 clusters. Training iterations were assessed by training substitute models for 1, 2, 3, and 4 iterations. The effect of the number of attack methods was tested with configurations of 1 (FD), 3 (FD, Hotflip, PSO), 5 (BAE, FD, Hotflip, PSO, TextBugger), and 7 (BAE, FD, Hotflip, PSO, TextBugger, LEAP, HQA) methods. Clustering methods were supplemented with K-means~\cite{lloyd1982least}, Spectral Clustering~\cite{von2007tutorial}, and BIRCH~\cite{zhang1996birch}. For embedding, we incorporated CLIP~\cite{radford2021learning} and one-hot encoding~\cite{rodriguez2018beyond}, in addition to the T5 pre-trained model. CLIP, trained on image-text pairs, uses a dataset distinct from typical text classification datasets, reducing overlap with the victim model’s data. One-hot encoding, a non-pretrained approach, relies solely on victim texts, avoiding external data interference.

 \textbf{Different cluster number.}
 As shown in Figure \ref{cluster-number-fig} (Appendix), the ASR decreases as the number of clusters increases. The ASR occurs with two clusters, validating our choice for VDBA.

\textbf{Different iteration.}
Table \ref{Different_iteration} presents the results of VDBA under varying training iterations. Increasing the number of iterations enhances ASR but leads to a significant decrease in similarity. Specifically, as the number of iterations increases from 1 to 4, ASR improves by at least 25\%, while Sim declines substantially. Notably, when the iterations increase from 3 to 4, the maximum Sim value drops to 0.913, with a particularly pronounced reduction. To balance attack effectiveness and the similarity of adversarial examples, we select 3 as the optimal number of training iterations.

\textbf{Different attack method number.}
Table \ref{Attack_method_number} (Appendix) summarizes the attack performance across different numbers of attack methods. Both ASR and Sim increase with the number of attack methods; however, the rate of improvement diminishes as the number of methods grows. Specifically, when the number of attack methods increases from 5 to 7, the improvements in ASR and Sim are negligible. Given that employing additional attack methods incurs higher computational costs, we select 5 attack methods as the optimal configuration for VDBA.

\textbf{Different clustering and embedding methods.}
The results presented in Figure \ref{cluster-vector} (Section \ref{section-cluster-vector} in Appendix) highlight the performance of different clustering and embedding methods, showing that neither method consistently outperforms the others. For clustering methods, the ASR and similarity metrics in subplots (a) and (b) indicate only minor differences in attack effectiveness, suggesting a limited and somewhat random impact of clustering on attack performance. In terms of embedding methods, as shown in subplots (c) and (d), CLIP and T5, both pre-trained models, exhibit comparable attack performance with similar ASR and similarity values. One-hot embedding, despite not requiring any training data, experiences only a small decrease in ASR (1.5\%) compared to the pre-trained models, with similarity largely unaffected. These findings demonstrate that VDBA remains effective with various embedding methods, including one-hot embedding, even without the need for pre-training data.

In summary,
using two clusters yields the highest ASR. Increasing attack iterations enhances ASR but decreases similarity, while adding attack methods improves both metrics until the effect stabilizes beyond five. Clustering and embedding methods introduce random variations, and VDBA remains effective with one-hot vectors, even for low-resource languages without pre-trained models.





\subsection{Realistic Scenario Attack}
\textbf{LLMs Attack.}
We evaluate VDBA's effectiveness in LLMs, including Qwen2~\cite{yang2024qwen2}, ChatGPT4o, and ChatGPT4omini~\cite{openai2023chatgpt4}, by comparing VDBA and transfer attacks. LLM agents are constructed using prompt learning (The detailed prompt is provided in Appendix \ref{prompt_word}).
In closed-source LLMs and prompt-based learning, where gradients and label probabilities are unavailable, our baseline shows that CT-GAT can still launch attacks. Thus, we compare CT-GAT and VDBA attack performance. Table \ref{llm} presents the experimental results, revealing that VDBA achieves an attack success rate exceeding 27\% across all datasets and SOTA LLMs, even with limited access to victim texts. These findings demonstrate that attackers can effectively target closed-source LLMs with minimal access to victim texts.

\begin{figure}
  \centering

  \includegraphics[width=0.47\textwidth]{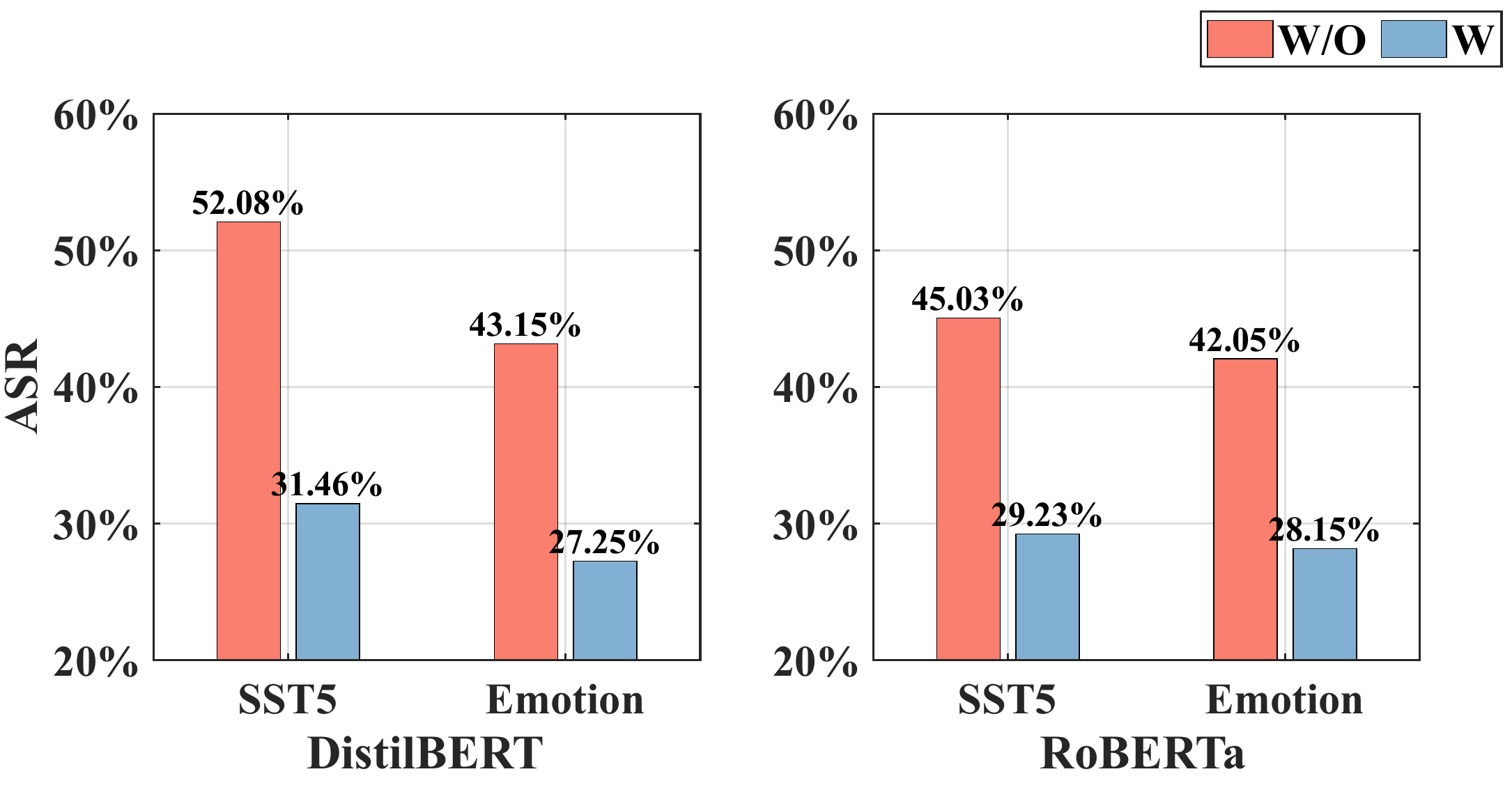} 
    \vspace{-10pt}
  \caption{The ASR of VDBA with and without the defense method. ``W" indicates the VDBA with the defense method. And ``W/O" indicates the VDBA without the defense method.
  \vspace{-10pt}
}\label{defense-results}

   \vspace{-10pt}
\end{figure}

\textbf{Attack results under defense method.}
Train-free preprocessing defenses and adversarial training are widely used defense mechanisms~\cite{qiu2019review}. We employ the preprocessing defense method proposed by \cite{wang2023punctuation}, which applies prompt learning on LLMs to mitigate adversarial text inputs.  Figure \ref{defense-results} shows ASR decreases significantly after applying this defense, although the attack remains partially effective. Additionally, the adversarial training results are presented in Appendix \ref{AT}. 

\textbf{Impact of victim model training data.}
We analyze the impact of victim model training data on attack performance, focusing on factors such as the number of class labels and  domain. We select 4 train datasets.
Table \ref{model_training_data} shows that when the training dataset and victim texts differ in attributes, ASR declines sharply. In contrast, when their attributes align, a higher number of class labels is associated with a smaller average distance and higher ASR.


\begin{table}[H]
\vspace{-15pt}
\centering
\caption{Impact of model training data on ASR of SST5 Dataset}
\label{model_training_data}
\resizebox{0.45\textwidth}{!}{
\begin{tabular}{@{}c|c|c|c@{}}
\toprule
Training Data        & \begin{tabular}[c]{@{}c@{}}Roberta \\ ASR(\%)↑\end{tabular} & \begin{tabular}[c]{@{}c@{}}Label \\ Number\end{tabular} & Domain    \\ \midrule \midrule
Financial PhraseBank & 33.76                                                    & 3                                                       & Sentiment \\
SST5         & 45.03                                                    & 5                                                       & Sentiment \\

Trec6                & 19.37                                                    & 6                                                       & Question  \\
Go Emotion           & 85.68                                                    & 28                                                      & Sentiment \\ \bottomrule
\end{tabular}
}
\vspace{-25pt}
\end{table}


\begin{table}[H]

\centering
\caption{The ASR(\%)↑ of LLMs.
}
\label{llm}
\resizebox{0.5\textwidth}{!}{

\begin{tabular}{@{}c|ccc|ccc@{}}
\toprule
Dataset      & \multicolumn{3}{c|}{SST5} & \multicolumn{3}{c}{Emotion} \\ \midrule
Victim Model & GPT4o & GPT4omini & Qwen2 & GPT4o  & GPT4omini  & Qwen2 \\ \midrule \midrule
CT-GAT       & 20.23 & 19.85     & 27.36 & 16.15  & 17.85      & 24.20 \\
VDBA          & 35.37 & 36.91     & 45.99 & 27.65  & 29.30       & 36.35 \\ \bottomrule

\end{tabular}

}
\vspace{-10pt}

\end{table}

\section{Conclusion }

We introduce a novel attack scenario, the VDBA, which is the first to show that text adversarial attacks can be performed without knowledge of the victim model, queries, or training data. Adversarial examples can be generated using only the victim texts. Both few-shot and one-shot scenarios demonstrate that adversarial examples can still be created with limited victim texts and minimal attribute knowledge. This victim texts-based scenario represents the most stringent attack condition, as it requires the least information, making it the most challenging. If an attack succeeds here, it is likely to succeed in other, less restrictive scenarios. VDBA’s potential extends to other modalities, functioning as a plug-and-play solution requiring only a publicly available pre-trained model. 

\section*{Impact Statement}
Our victim data-based attack is an adversarial attack based on the victim's texts. Our goal is to advance the fields of machine learning and textual adversarial attacks. Our experiments are conducted on public datasets. Our work does not involve any ethical considerations or have immediate societal consequences.


\bibliography{example_paper}
\bibliographystyle{icml2025}

\newpage
\appendix
\onecolumn
This appendix includes our supplementary materials as follows:

- More results on additional datasets in Section \ref{more-result}

- More details of baselines in Section \ref{baseline}

- More detail of dataset in Section \ref{datasets}

- More detail of evaluate metrics in Section \ref{em}

- More detail of substitute model in Section \ref{surrogate-model}

- More mathematical proofs of Theorem \ref{max_prob1} in Section \ref{proof_max_prob1} 

- More mathematical proofs of Theorem \ref{max_prob2} in Section \ref{proof_max_prob2}

- The Detailed results of  clustering and embedding methods in 
section \ref{section-cluster-vector}

- The Results of Adversarial Training in Section \ref{AT}

- Prompt of Emotion and SST5 Datasets in LLMs Attack in 
section \ref{prompt_word}

- The Unlimited Baselines Results in Section \ref{unlimited}

\section{Further Experimental Results on Additional Datasets}
\label{more-result}
The results for the AG News and TREC6 datasets are presented in Table \ref{More-data-results}. Consistent with the results from the SST5 and Emotion datasets, VDBA achieves state-of-the-art (SOTA) performance in terms of ASR and Queries metrics while maintaining substantial similarity. Although our similarity scores do not achieve SOTA performance, they remain close to it, as shown in Table \ref{More-data-results}. Given that VDBA operates without access to any information about the victim model or its labels and relies solely on the victim texts, this trade-off is justified. The reduction in similarity enables VDBA to generate adversarial examples under highly stringent conditions while achieving a significant ASR. We consider this trade-off worthwhile, as it requires sacrificing only a minimal amount of similarity.
\subsection{AG News Dataset}
The AG News dataset is a widely used benchmark for news classification in natural language processing (NLP). It consists of 127,600 news articles in training set  7,600 in  test set, which can be categorized into four classes: World, Sports, Business, and Science/Technology. Each sample includes a title and description, making it suitable for text classification, topic modeling, and deep learning benchmarking. The dataset has been extensively used to evaluate machine learning models, including CNNs, RNNs, and transformer-based architectures such as BERT and RoBERTa. Common applications include domain adaptation, few-shot learning, and evaluating model generalization. Due to its structured labeling and diverse category distribution, AG News remains a standard dataset for assessing text classification performance in NLP research.

\subsection{TREC6 Dataset}
The TREC6 dataset is a widely used benchmark for question classification in natural language processing (NLP), originally introduced as part of the Text Retrieval Conference (TREC). It comprises 5,952 labeled questions, with 5,452 samples for training and 500 for testing, categorized into six broad classes: Abbreviation (ABBR), Entity (ENTY), Description (DESC), Human (HUM), Location (LOC), and Numeric (NUM). The dataset is extensively used to evaluate machine learning and deep learning models, including CNNs, LSTMs, and Transformer-based architectures such as BERT and RoBERTa. Common applications include text classification, question answering, few-shot learning, and benchmarking NLP models. Due to its well-defined categories and structured format, TREC-6 remains a standard dataset for assessing question classification performance in NLP research.
\begin{table*}[t]
\caption{The attack performance of VDBA and other attack methods on TREC6 and AG News Datasets.  For each metric, the best method is highlighted in \textbf{bold} and the runner-up is \underline{underlined}.}\label{More-data-results}

\resizebox{1\textwidth}{!}{
\centering
\begin{tabular}{@{}ccccccc|cccccc@{}}
\toprule
\multirow{3}{*}{Method} & \multicolumn{6}{c|}{TREC6}                                        & \multicolumn{6}{c}{AG News}                                     \\ \cmidrule(l){2-13} 
                        & \multicolumn{3}{c}{DistilBERT}    & \multicolumn{3}{c|}{RoBERTa} & \multicolumn{3}{c}{DistilBERT}    & \multicolumn{3}{c}{RoBERTa} \\ \cmidrule(l){2-13} 
                        & ASR(\%) $\uparrow$ & Sim $\uparrow$ & Queries $\downarrow$               & ASR(\%) $\uparrow$ & Sim $\uparrow$ & Queries $\downarrow$               & ASR(\%) $\uparrow$ & Sim $\uparrow$ & Queries $\downarrow$               & ASR(\%) $\uparrow$ & Sim $\uparrow$ & Queries $\downarrow$               \\ \midrule \midrule
 Bae& 22.40      &     0.761               & \multicolumn{1}{c|}{4584             
} &        21.20      &        0.75                &            4540             
& 21.43      &     0.808               & \multicolumn{1}{c|}{95288            } &        25.64      &        0.776               &           87063            
\\
FD                      & 27.80      &     0.871               & \multicolumn{1}{c|}{9686             
} &        31.40      &        0.873               &            9446             
&

38.25      &     0.866               & \multicolumn{1}{c|}{228182           } &        \underline{39.78}&        0.865               &           80864            
\\
  Hotflip& \underline{38.80}&     0.899               & \multicolumn{1}{c|}{\underline{3669}} &        37.40      &        0.9                 &            3605             
& 38.25      &     0.842               & \multicolumn{1}{c|}{77809            } &        37.20      &        0.828               &           61403            
\\
PSO                     & 35.00      &     0.554               & \multicolumn{1}{c|}{3233             
} &        34.20      &        0.927               &            3162             
&

34.33      &     0.88                & \multicolumn{1}{c|}{63134            } &        36.57      &        0.822               &           56569            
\\
 TextBug                 & 37.80      &     0.942               & \multicolumn{1}{c|}{7771             
} &        \underline{39.60}&        \underline{0.957}&            7607             
& \underline{43.71}&     0.881               & \multicolumn{1}{c|}{142530           } &        38.68      &        0.903               &           138905           
\\
Leap                    & 38.66      &     0.886               & \multicolumn{1}{c|}{13700            
} &        39.79      &        0.918               &            13550            
&

26.64      &     0.896               & \multicolumn{1}{c|}{267596           } &        27.39      &        0.92                &           258020           
\\
CT-GAT                    & 
10.40      &     0.959               & \multicolumn{1}{c|}{5994             
} &        9.60       &        0.980                &            \underline{6009}& 20.36      &     \underline{0.918}& \multicolumn{1}{c|}{\underline{120825}} &        34.13      &        \textbf{0.971}&           \underline{111028}\\
 HQA                     & 34.60      &     \underline{0.944}& \multicolumn{1}{c|}{13297            
} &        36.00      &        0.946               &            13642            
&

30.01      &     0.931               & \multicolumn{1}{c|}{220970           } &        
34.86      &        \underline{0.957}&           211105           
\\
\midrule \rowcolor{orange!50}  \textbf{VDBA}                     & 
\textbf{42.80}&   0.933               & \multicolumn{1}{c|}{\textbf{0}                } &        \textbf{43.60}&       0.935               &            \textbf{0}                & \textbf{46.33}&     0.911               & \multicolumn{1}{c|}{\textbf{0}                } &       \textbf{41.51}&        0.932               &          \textbf{0}                \\ \bottomrule
\end{tabular}

}
\vspace{-10pt}
\end{table*}

\section{The Details of Baselines Methods}\label{baseline}
In this section, we present the details of the attack methods we used, including Bae, FD, Hotflip, PSO and TextBug.

\textbf{Bae}: Bae (BERT-based Adversarial Examples) is an advanced attack method utilizing the BERT pre-trained model and prompt learning. This approach involves systematically replacing words in the input text to create adversarial examples, thereby testing the robustness of natural language processing models.

\textbf{FD}: The FD (Frequency Domain) attack method replaces words with synonyms based on the gradient descent optimization of the victim model. This technique aims to subtly alter the input text to generate adversarial examples while maintaining semantic coherence.

\textbf{Hotflip~\cite{ebrahimi2018Hotflip}}: Hotflip~\cite{ebrahimi2018Hotflip} is a text white-box attack method. It iteratively replaces individual words based on their calculated importance. The importance of each word is determined by the magnitude of the gradient in the victim model, allowing for precise identification of the most impactful words to alter.

\textbf{PSO}: PSO is a text soft-label black-box attack method. It employs a versatile metaheuristic approach to optimize the search space for generating adversarial examples, thereby enabling the victim model to produce varied outputs. This method leverages the sememe-based representations of words to effectively navigate and perturb the input text.

\textbf{TextBug}: TextBug is an attack framework capable of generating adversarial text for real-world applications, functioning in both white-box and black-box environments. It perturbs words based on their significance, using the Jacobian matrix to identify key words in white-box settings and a scoring function in black-box settings. It is effective and efficient, maintaining the original utility of the text while achieving high success rates in evading state-of-the-art NLP systems.

\section{The Details of Datasets}\label{datasets}
In this section, we present the details of the datasets we used, including Tomatoes, Emotion, Go-emotion, and SST5 datasets. The resuls is presented in Table \ref{stat}.
The \textbf{Emotion} dataset, which encompasses six distinct Emotions, is derived from Twitter messages. \textbf{SST5} The SST-5 dataset, a sentiment analysis resource comprising five categories, originates from movie reviews. 

\begin{table*}[h]
\centering
\caption{The statistics of datasets.}
\label{stat}
\resizebox{\textwidth}{!}{%
\begin{tabular}{@{}c|ccccc@{}}
\toprule
Dataset         & Train & Test & Type      & Number of labels & Labels name       \\ \midrule
Tomatoes & 8530  & 1066 & Sentiment & 2                & Positive, Negtive \\
SST5       & 8544  & 2210 & Sentiment & 5  & Very positive, Positive, Neutral, Negative, Very negative \\
Go-emotions & 43410 & 5427 & Sentiment & 28 & Admiration, Amusement, Anger, ... , Surprise, Neutral     \\
Emotion    & 16000 & 2000 & Sentiment & 6  & Sadness, Joy, Love, Anger, Fear, Surprise                 \\ \bottomrule
\end{tabular}%
}
\end{table*}

\section{The Details of  Evaluation Metrics}\label{em}
\textbf{Attack Success Rate:} 
Attack Success Rate is calculated as the ratio of the number of success attack adversarial examples to the total number of all adversarial examples. The higher ASR signifies the better attack method.

\textbf{Semantic Similarity:} Semantic Similarity is assessed by computing the mean similarity between the perturbed texts and the original texts. An elevated Semantic Similarity implies a more potent attack strategy.

\textbf{Number of Queries:} This metric denotes the quantity of Queries that attackers direct towards the victim model. A reduced query count suggests a more efficient attack method.

\section{The Details of  Substitute Model}\label{surrogate-model}
The substitute model is based on a transformer architecture with 12 hidden layers of size 768, and a dropout probability of 0.1. The model is trained using the AdamW optimizer with a batch size of 32, a learning rate of 0.00005, and over 2 epochs. The model consists of 12 transformer blocks, each containing 768 hidden units and 12 self-attention heads. Each transformer block includes specific substructures as detailed below.

\begin{itemize}
    \item \textbf{Self-Attention Layer:} The hidden size of the self-attention layer is $768$.

    \item \textbf{Position-wise Feed-Forward Networks:} This network first maps the output of the attention layer to a $3072$-dimensional feature space through a fully connected layer, then applies a ReLU activation function for non-linear activation, and finally maps the $3072$-dimensional feature space back to a $768$-dimensional feature space through a second fully connected layer.

    \item \textbf{Layer Normalization and Residual Connection:}
    \begin{itemize}
        \item \textbf{Layer Normalization:} Applied to the output of each sub-layer to stabilize the training process.
        \item \textbf{Residual Connection:} Adds the normalized output to the input of the sub-layer.
    \end{itemize}
\end{itemize}

\section{The Proof of Theorem \ref{max_prob1}}\label{proof_max_prob1}
\begin{proof}

We aim to prove that as \( m \to \infty \), the probability that the maximum of the success probabilities \( \{ p_1^s, p_2^s, \dots, p_m^s \} \) exceeds any fixed success probability \( p_i^s \) approaches 1, for \( p_i^s < 1 \). We will also prove that the maximum of the success probabilities exceeds the maximum of a smaller subset as \( m \to \infty \).

\textbf{1. Probability of Maximum Exceeding \( p_i^s \)}

We start by analyzing the probability that the maximum of the \( m \) success probabilities exceeds \( p_i^s \). Let \( \max\{ p_1^s, p_2^s, \dots, p_m^s \} \) represent the maximum success probability among the \( m \) models. We seek to find:

\begin{equation}
\Pr\left( \max\{ p_1^s, p_2^s, \dots, p_m^s \} > p_i^s \right).
\end{equation}

This can be rewritten using the complement rule. The event that \( \max\{ p_1^s, p_2^s, \dots, p_m^s \} > p_i^s \) is the complement of the event where all \( p_j^s \) for \( j = 1, 2, \dots, m \) are less than or equal to \( p_i^s \). Hence:

\begin{equation}
\Pr\left( \max\{ p_1^s, p_2^s, \dots, p_m^s \} > p_i^s \right) = 1 - \Pr\left( p_1^s \leq p_i^s, p_2^s \leq p_i^s, \dots, p_m^s \leq p_i^s \right).
\end{equation}

Since the success probabilities \( p_1^s, p_2^s, \dots, p_m^s \) are assumed to be independent, the probability that all \( m \) success probabilities are less than or equal to \( p_i^s \) is the product of the individual probabilities:

\begin{equation}
\Pr\left( p_1^s \leq p_i^s, p_2^s \leq p_i^s, \dots, p_m^s \leq p_i^s \right) = \prod_{j=1}^{m} \Pr(p_j^s \leq p_i^s) = \prod_{j=1}^{m} F(p_i^s) = F(p_i^s)^m.
\end{equation}

Thus, the probability that the maximum exceeds \( p_i^s \) is:

\begin{equation}
\Pr\left( \max\{ p_1^s, p_2^s, \dots, p_m^s \} > p_i^s \right) = 1 - F(p_i^s)^m.
\end{equation}

\textbf{2. Behavior as \( m \to \infty \)}

Now, consider the behavior of this probability as \( m \to \infty \). If \( F(p_i^s) < 1 \), then as \( m \) increases, \( F(p_i^s)^m \) tends to 0. Therefore:

\begin{equation}
\lim_{m \to \infty} \Pr\left( \max\{ p_1^s, p_2^s, \dots, p_m^s \} > p_i^s \right) = 1 - 0 = 1.
\end{equation}

Thus, as the number of models \( m \) grows, the probability that at least one model has a success probability greater than \( p_i^s \) approaches 1. This proves the first part of the theorem.

\textbf{3. Probability of Maximum Exceeding the Maximum of a Subset}

Next, we prove the second part of the theorem, which states that as \( m \to \infty \), the probability that the maximum of the \( m \) success probabilities exceeds the maximum of a smaller subset of \( m_1 \) success probabilities also approaches 1.

Let \( p_{max} = \max\{ p_1^s, p_2^s, \dots, p_{m_1}^s \} \) be the maximum of the first \( m_1 \) success probabilities. We seek to find the probability:

\begin{equation}
\Pr\left( \max\{ p_1^s, p_2^s, \dots, p_m^s \} > p_{max} \right).
\end{equation}

Again, using the complement rule, this is:

\begin{equation}
\Pr\left( \max\{ p_1^s, p_2^s, \dots, p_m^s \} > p_{max} \right) = 1 - \Pr\left( p_1^s \leq p_{max}, p_2^s \leq p_{max}, \dots, p_m^s \leq p_{max} \right).
\end{equation}

Since the success probabilities are independent, the probability that all \( m \) success probabilities are less than or equal to \( p_{max} \) is:

\begin{equation}
\Pr\left( p_1^s \leq p_{max}, p_2^s \leq p_{max}, \dots, p_m^s \leq p_{max} \right) = \prod_{j=1}^{m} F(p_{max}) = F(p_{max})^m.
\end{equation}

Thus, the probability that the maximum exceeds \( p_{max} \) is:

\begin{equation}
\Pr\left( \max\{ p_1^s, p_2^s, \dots, p_m^s \} > p_{max} \right) = 1 - F(p_{max})^m.
\end{equation}

As \( m \to \infty \), if \( F(p_{max}) < 1 \), then \( F(p_{max})^m \to 0 \), and hence:

\begin{equation}
\lim_{m \to \infty} \Pr\left( \max\{ p_1^s, p_2^s, \dots, p_m^s \} > p_{max} \right) = 1.
\end{equation}

This completes the proof of the second part of the theorem.

\textbf{Conclusion:} We have shown that as the number of substitute models \( m \) increases, the probability that the maximum of the success probabilities exceeds any fixed \( p_i^s \) or even the maximum of a smaller subset of success probabilities approaches 1. Therefore, the theorem holds.

\end{proof}

\section{The Proof of Theorem \ref{max_prob2}}\label{proof_max_prob2}
\begin{proof}
\textbf{Step 1: Monotonicity of $\mu_w$.}

By definition,
\[
\mu_w \;=\; \mathbb{E}\biggl[\sum_{i=1}^w X_i\biggr] 
\;=\; \sum_{i=1}^w \mathbb{E}[X_i]
\;=\; \sum_{i=1}^w p_i^{\text{sim}}.
\]
Since $p_i^{\text{sim}} \ge 0$ for each $i$, adding one more method $M_{w+1}$ (with $p_{w+1}^{\text{sim}} \ge 0$) yields
\[
\mu_{w+1} 
\;=\; \mu_w + p_{w+1}^{\text{sim}}
\;\ge\; \mu_w.
\]
Hence, $\mu_w$ is a non-decreasing function of $w$. Moreover, if $p_{w+1}^{\text{sim}} > 0$, we strictly have $\mu_{w+1} > \mu_w$.

\medskip
\textbf{Step 2: Monotonicity of $P^{\text{sim}}_{(w)}$.}

Define
\[
P^{\text{sim}}_{(w)} 
\;=\; \mathbb{P}\!\Bigl(\bigcup_{i=1}^w \{X_i = 1\}\Bigr).
\]
Under the assumption of (approximate) independence among the methods $\{ M_i \}$, we have
\[
P^{\text{sim}}_{(w)} 
\;=\; 1 \;-\; \mathbb{P}\!\Bigl(\bigcap_{i=1}^w \{X_i = 0\}\Bigr)
\;=\; 1 \;-\; \prod_{i=1}^w (1 - p_i^{\text{sim}}).
\]
Since each $p_i^{\text{sim}} \ge 0$, adding another method $M_{w+1}$ with $p_{w+1}^{\text{sim}} \ge 0$ cannot decrease the product term, and thus cannot decrease $1 - \prod_{i=1}^w (1 - p_i^{\text{sim}})$. Therefore, $P^{\text{sim}}_{(w)}$ is non-decreasing in $w$. If $p_{w+1}^{\text{sim}} > 0$, then $P^{\text{sim}}_{(w+1)} > P^{\text{sim}}_{(w)}$.

\medskip
\textbf{Step 3: Limit as $w \to \infty$.}

If infinitely many $p_i^{\text{sim}}$ are strictly greater than 0 and the $X_i$'s are (approximately) independent, then
\[
\prod_{i=1}^\infty \bigl(1 - p_i^{\text{sim}}\bigr)
\;=\; 0,
\]
because each factor $(1 - p_i^{\text{sim}})$ is less than 1 for infinitely many $i$, causing the infinite product to tend to 0. Consequently,
\[
\lim_{w \to \infty} P^{\text{sim}}_{(w)}
\;=\; 1 
\quad \text{and} \quad
\mu_w \to \infty \; \text{(if } p_i^{\text{sim}} \text{ sums to infinity)}.
\]
Thus, $P^{\text{sim}}_{(w)}$ approaches 1 as $w \to \infty$, satisfying
\[
\lim_{w \to \infty} P^{\text{sim}}_{(w)} = 1,
\]
and for any finite $w > h$, $P^{\text{sim}}_{(w)} > P^{\text{sim}}_{(h)}$ and $\mu_w > \mu_h$, as required.
\end{proof}

\section{The Detailed results of  clustering and embedding methods}\label{section-cluster-vector}
\begin{figure*}[h]
  \centering
  \includegraphics[width=0.81\textwidth]{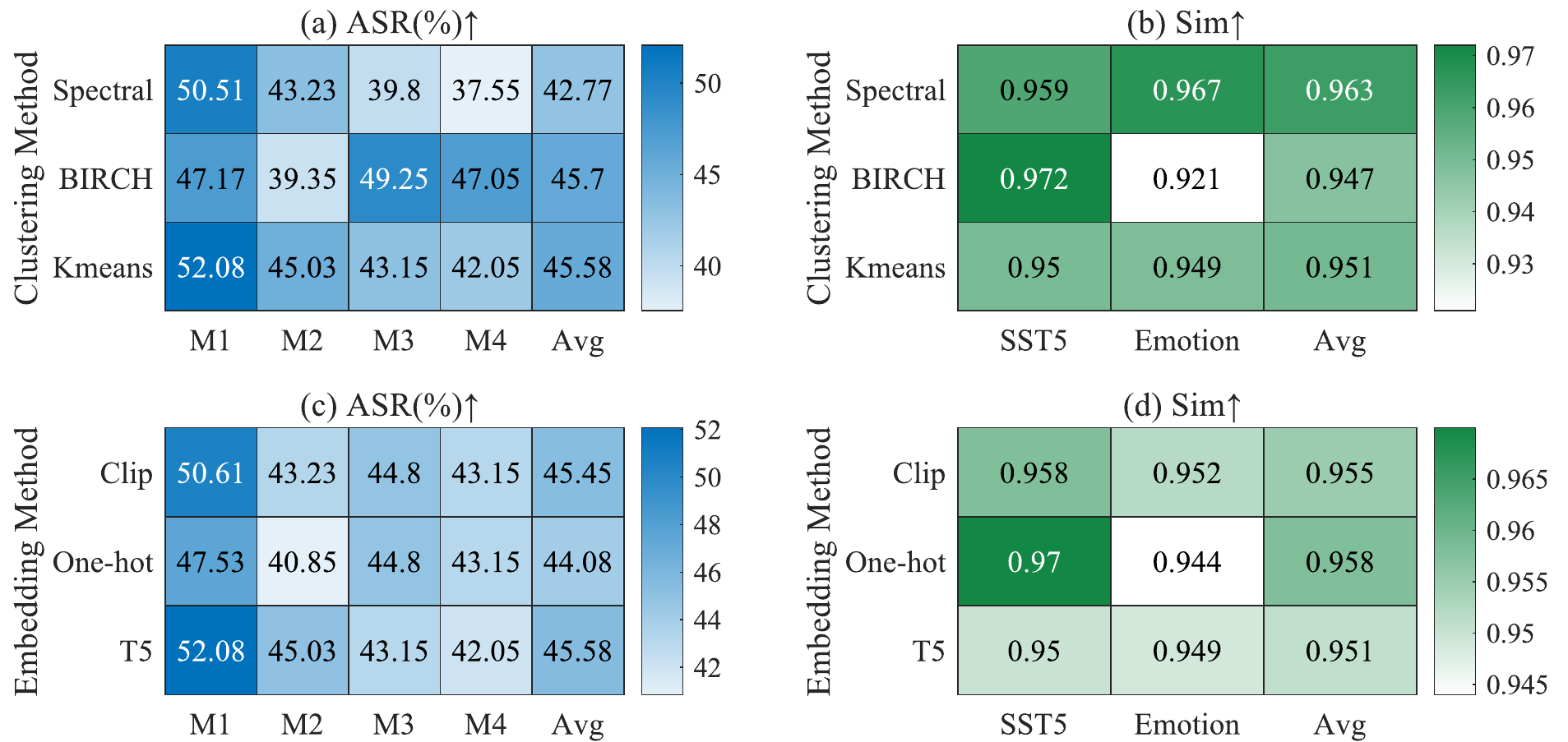} 
 
  \caption{Experiment results under different clustering vectorization method. M1 and M2 represent DistilBERT and RoBERTa models for SST5, while M3 and M4 represent DistilBERT and RoBERTa models for Emotion. ``Avg'' denotes the mean score of ASR and similarity. \textbf{
The impact of clustering and vectorization methods on the attack effectiveness is random.}}\label{cluster-vector}

\end{figure*}

\section{The ASR for different cluster numbers}
\begin{figure}
  \centering
  \includegraphics[width=0.33\textwidth]{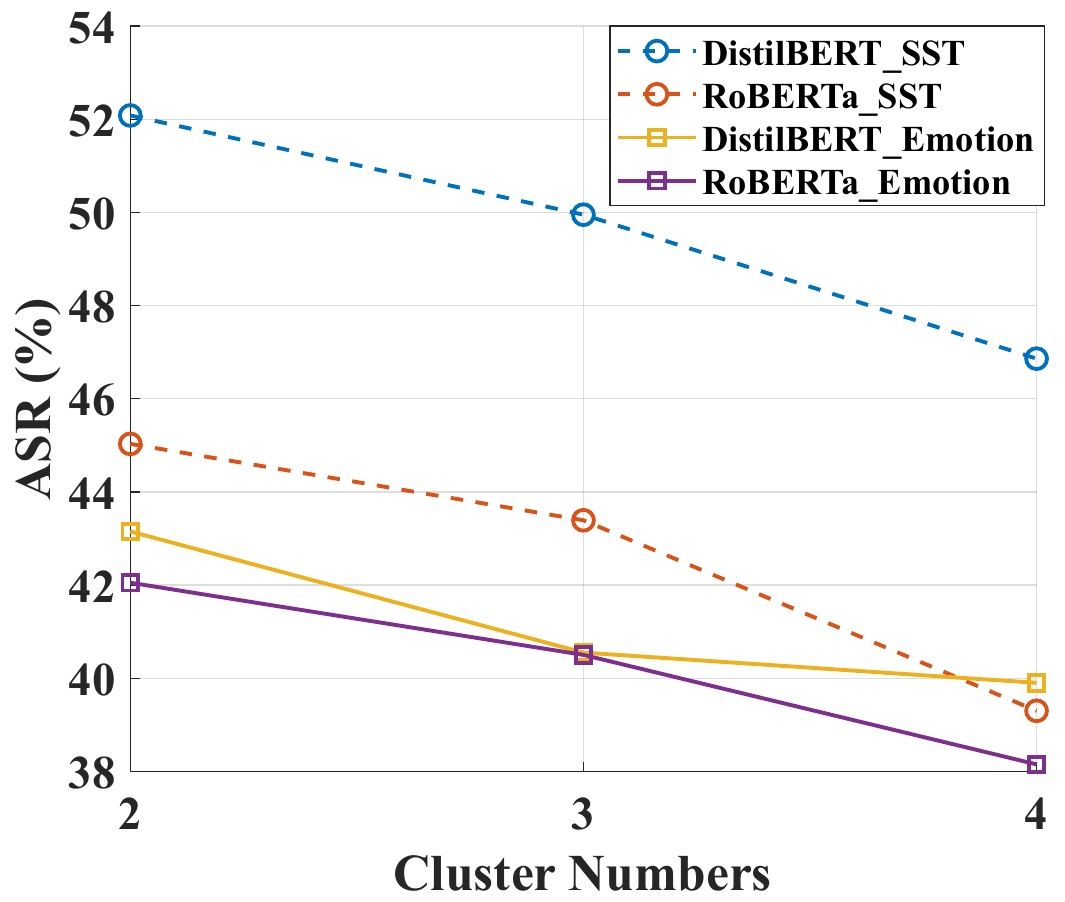} 
  \caption{The ASR for different cluster numbers. \textbf{Fewer clusters imply better ASR.}}
  \label{cluster-number-fig}

\end{figure}

The results presented in Figure \ref{cluster-vector}  provide a comprehensive analysis of the performance across different clustering and embedding methods. The data reveal that no single method consistently outperforms the others across all metrics. Specifically, in the case of clustering methods, the Attack Success Rate (ASR) and similarity metrics, as depicted in subplots (a) and (b), show only marginal differences in attack effectiveness. These minor variations suggest that the choice of clustering method has a limited and somewhat random impact on the overall attack performance, indicating that clustering may not play a significant role in optimizing adversarial outcomes.

Regarding embedding methods, the performance metrics displayed in subplots (c) and (d) demonstrate that the pre-trained models CLIP and T5 yield comparable attack effectiveness, with similar ASR and similarity values. This suggests that these models, despite their distinct architectures and training strategies, offer equivalent levels of adversarial robustness. In contrast, the one-hot embedding method, despite its simplicity and lack of pre-training data, shows only a slight decrease in ASR (approximately 1.5\%) compared to the pre-trained models, with similarity values remaining largely unchanged. This finding is particularly noteworthy, as it suggests that VDBA retains its effectiveness even when utilizing embeddings that do not require extensive training datasets. Therefore, these results highlight the versatility and robustness of VDBA across a range of embedding methods, including one-hot encoding, making it a viable approach even in scenarios where pre-trained models or large training datasets are unavailable.

\section{The Attack Performance of VDBA under Varying Iterations of the Hierarchical Substitute Model Design }
\begin{table}[t]
\centering
\caption{Attack performance of VDBA under varying iterations of the hierarchical substitute model design. \textbf{Increasing the number of iterations improves ASR but reduces similarity.}}
\label{Different_iteration}

\begin{tabular}{@{}c|c|ccc@{}}
\toprule
Data                     & Iteration & \begin{tabular}[c]{@{}c@{}}Roberta \\ ASR(\%)↑\end{tabular} & \begin{tabular}[c]{@{}c@{}}Distilbert \\ ASR(\%)↑\end{tabular} & Sim. ↑ \\ \midrule \midrule
\multirow{4}{*}{SST5}    & 1         & 32.29                                                       & 40.90                                                          & 0.962  \\
                         & 2         & 37.68                                                       & 45.42                                                          & 0.954  \\
                         & 3         & 45.03                                                       & 52.08                                                          & 0.950  \\
                         & 4         & 58.42                                                       & 64.49                                                          & 0.916  \\ \midrule
\multirow{4}{*}{Emotion} & 1         & 30.65                                                       & 28.55                                                          & 0.978  \\
                         & 2         & 35.20                                                       & 33.95                                                          & 0.966  \\
                         & 3         & 42.05                                                       & 43.15                                                          & 0.949  \\
                         & 4         & 57.85                                                       & 61.60                                                          & 0.913  \\ \bottomrule
\end{tabular}
\end{table}

\section{Attack Performance of VDBA  under Varying Numbers of Attack Methods}
\begin{table}[t]

\caption{Attack performance of VDBA is evaluated under varying numbers of attack methods in diverse adversarial example generation. Both ASR and Sim increase with the number of attack methods; however, the growth rate of ASR and Sim diminishes as the number of attack methods further increases.}
\label{Attack_method_number}

\centering
\begin{tabular}{@{}c|c|ccc@{}}

\toprule
Data                     & \begin{tabular}[c]{@{}c@{}}Method \\ Number\end{tabular} & \begin{tabular}[c]{@{}c@{}}Roberta \\ ASR(\%)↑\end{tabular} & \begin{tabular}[c]{@{}c@{}}Distilbert \\ ASR(\%)↑\end{tabular} & Sim. ↑ \\ \midrule \midrule \midrule
\multirow{4}{*}{SST5}    & 1                                                        & 23.71                                                       & 31.27                                                          & 0.916  \\
                         & 3                                                        & 33.23                                                       & 39.29                                                          & 0.925  \\
                         & 5                                                        & 45.03                                                       & 52.08                                                          & 0.950  \\
                         & 7                                                        & 50.46                                                       & 56.44                                                          & 0.957  \\ \midrule
\multirow{4}{*}{Emotion} & 1                                                        & 32.30                                                       & 31.65                                                          & 0.923  \\
                         & 3                                                        & 37.35                                                       & 38.05                                                          & 0.936  \\
                         & 5                                                        & 42.05                                                       & 43.15                                                          & 0.949  \\
                         & 7                                                        & 43.50                                                       & 46.55                                                          & 0.954  \\ \bottomrule
\end{tabular}

\end{table}

\section{The Results of Adversarial Training}\label{AT}
In the original study, the victim models are well-trained, and we examine the defensive strategies employed by language correctors. In this work, we investigate the impact of adversarial training on attack efficacy. Classifiers are trained on the SST5 and Emotion datasets using the pre-trained BERT model. Subsequently, adversarial training is implemented by generating adversarial examples using the BAE, FD, HotFlip, PSO, and TextBug attack methods. These adversarial examples are integrated with the original training data to form an augmented training set, on which the model is then trained to accurately classify both original and adversarial samples. Before adversarial training, the Attack Success Rate (ASR) of the Emotion and SST5 datasets is 46.20\% and 49.81\%, respectively. After adversarial training, the ASR for these datasets decreases to 23.85\% and 22.27\%, respectively. These results indicate that VDBA remains effective, achieving considerable success even under adversarial training.

\section{Prompt of Emotion and SST5 Datasets in LLMs Attack}\label{prompt_word}

\subsection{Prompt of  Emotion Datasets in LLMs Attack}
**Prompt of Emotion dataset:**  
\begin{quote}
``The objective is to predict the label of the provided text. It is sufficient to supply the label alone. The labels encompass `Anger', `Fear', `Joy', `Love', `Sadness', and `Surprise', excluding any other labels."
\end{quote}
\subsection{Prompt of  SST5 Datasets in LLMs Attack}
\textbf{Prompt of SST5 dataset:} 
\begin{quote}
``The objective is to predict the label of the provided text. It is sufficient to supply the label alone. The labels encompass `Very negtive', `Negtive', `Neutral,', `Positive', and `Very positive', excluding any other labels."
\end{quote}

\end{document}